\documentclass[runningheads]{llncs}
\usepackage{graphicx}
\usepackage{tikz}
\usepackage{comment}
\usepackage{amsmath,amssymb} 
\usepackage{color}
\usepackage[accsupp]{axessibility}  

\usepackage[width=122mm,left=12mm,paperwidth=146mm,height=193mm,top=12mm,paperheight=217mm]{geometry}
\usepackage{booktabs}
\usepackage{enumitem}
\usepackage{multicol}
\usepackage{multirow}
\usepackage{comment}
\usepackage{soul}
\usepackage{microtype}

\newenvironment{tight_itemize}{
\begin{itemize}[leftmargin=15pt,nosep]
  \setlength{\topsep}{0pt}
  \setlength{\itemsep}{0pt}
  \setlength{\parskip}{0pt}
  \setlength{\parsep}{0pt}
}{\end{itemize}}

\newcommand{\mc}[1]{\textcolor{black}{{#1}}}
\newcommand{\zl}[1]{\textcolor{black}{{#1}}}
\newcommand{\zlnew}[1]{\textcolor{black}{{#1}}}

\usepackage[pagebackref,breaklinks,colorlinks]{hyperref}

\usepackage[capitalize]{cleveref}
\crefname{section}{Sec.}{Secs.}
\Crefname{section}{Section}{Sections}
\Crefname{table}{Table}{Tables}
\crefname{table}{Tab.}{Tabs.}

\def\ECCVSubNumber{1276}  

\begin{document}
\pagestyle{headings}
\mainmatter

\title{Physically-Based Editing of Indoor Scene Lighting from a Single Image} 

%
\author{Zhengqin Li\inst{1} \and
Jia Shi\inst{1,3} \and
Sai Bi \inst{1,2} \and
Rui Zhu\inst{2} \and 
Kalyan Sunkavalli\inst{2} \and 
Milo\v{s} Ha\v{s}an\inst{2} \and
Zexiang Xu\inst{2} \and 
Ravi Ramamoorthi \inst{1} \and
Manmohan Chandraker \inst{1}
}
\authorrunning{Z. Li et al.}

\institute{UC San Diego \and
Adobe Research
\and
Carnegie Mellon University}

\maketitle

\begin{abstract}
We present a method to edit complex indoor lighting from a single image with its predicted depth and light source segmentation masks. This is an extremely challenging problem that requires modeling complex light transport, and disentangling HDR lighting from material and geometry with only a partial LDR observation of the scene. We tackle this problem using two novel components: 1) a holistic scene reconstruction method that estimates scene reflectance and parametric 3D lighting, and 2) a neural rendering framework that re-renders the scene from our predictions. We use physically-based indoor light representations that allow for intuitive editing, and infer both visible and invisible light sources. Our neural rendering framework combines physically-based direct illumination and shadow rendering with deep networks to approximate global illumination. It can capture challenging lighting effects, such as soft shadows, directional lighting, specular materials, and interreflections. Previous single image inverse rendering methods usually entangle scene lighting and geometry and only support applications like object insertion. Instead, by combining parametric 3D lighting estimation with neural scene rendering, we demonstrate the first automatic method to achieve full scene relighting, including light source insertion, removal, and replacement, from a single image.  All source code and data will be publicly released.
\end{abstract}

\begin{figure}[t]
\centering
 \includegraphics[width=\textwidth]{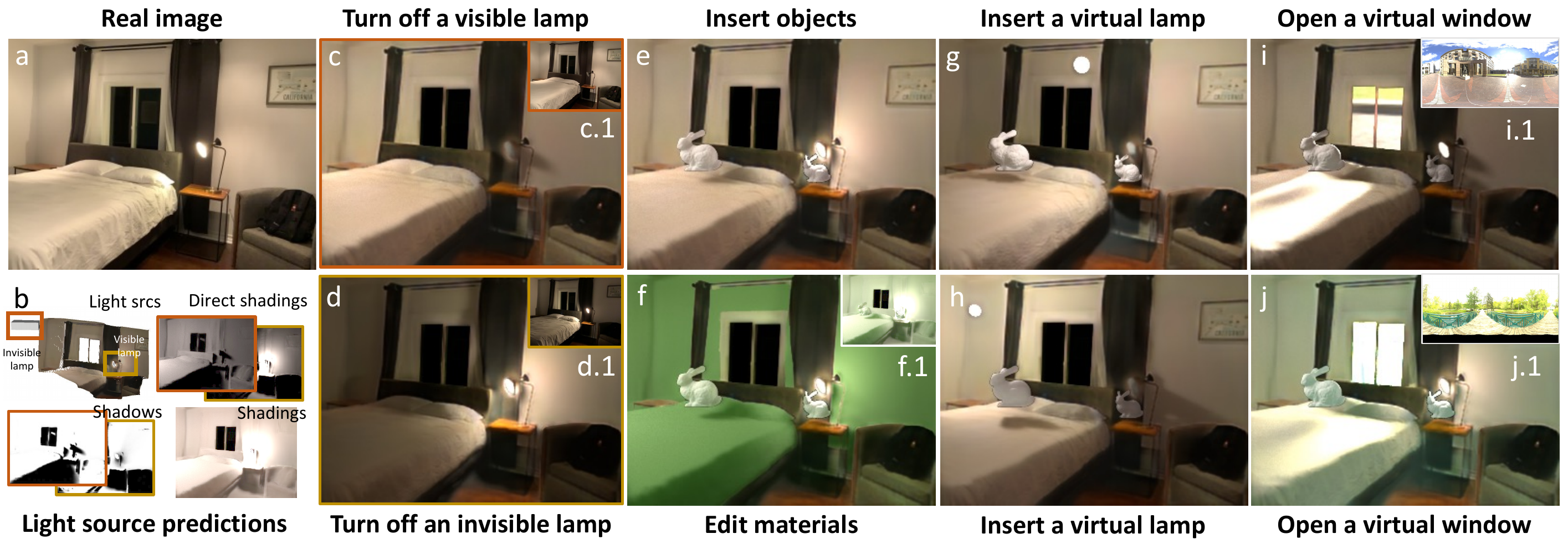}
 \vspace{-0.8cm}
\caption{We present the first method \mc{for globally consistent editing of} indoor scene lighting from a single LDR image. Given the input (a), our framework first estimates physically-based light source parameters, for both visible and invisible lights, and then renders their direct contributions and interreflections through a neural rendering framework (b). Our framework can turn off visible and invisible light sources (c and d) with results that closely match the ground truths (c.1 and d.1). It can insert virtual objects at arbitrary locations (e) with consistent changes of highlight and shadow and edit materials with color bleeding being correctly rendered, as shown in the rendered image (f) and shading (f.1). It can also insert virtual lamps (g and h) and open a virtual window (i and j) to \zl{let sunlight} (i.1 and j.1) shine into the room.} 
\label{fig:teaser}
\vspace{-0.6cm}
\end{figure}

\vspace{-0.8cm}
\section{Introduction}
\label{sec:introduction}

Light sources of various shapes, colors and types, such as lamps and windows, play an important role in determining indoor scene appearances. Their influence leads to several interesting phenomena such as light shafts through an open window on a sunlit day, highlights on specular surfaces due to incandescent lamps, interreflections from colored walls, or shadows cast by furniture in the room. Correctly attributing those effects to individual visible or invisible light sources in a single image enables abilities for photorealistic augmented reality that have previously been intractable — virtual furniture insertion under varying illuminations with consistent highlights and shadows, virtual try-on of wall paints with accurate global interreflections, or morphing a room under fluorescent lights into one reflecting the sunrise through a window (Fig.~\ref{fig:teaser}).

\begin{figure}[!!t]
\centering
\begin{minipage}[c]{0.6\linewidth}
\includegraphics[width=\columnwidth]{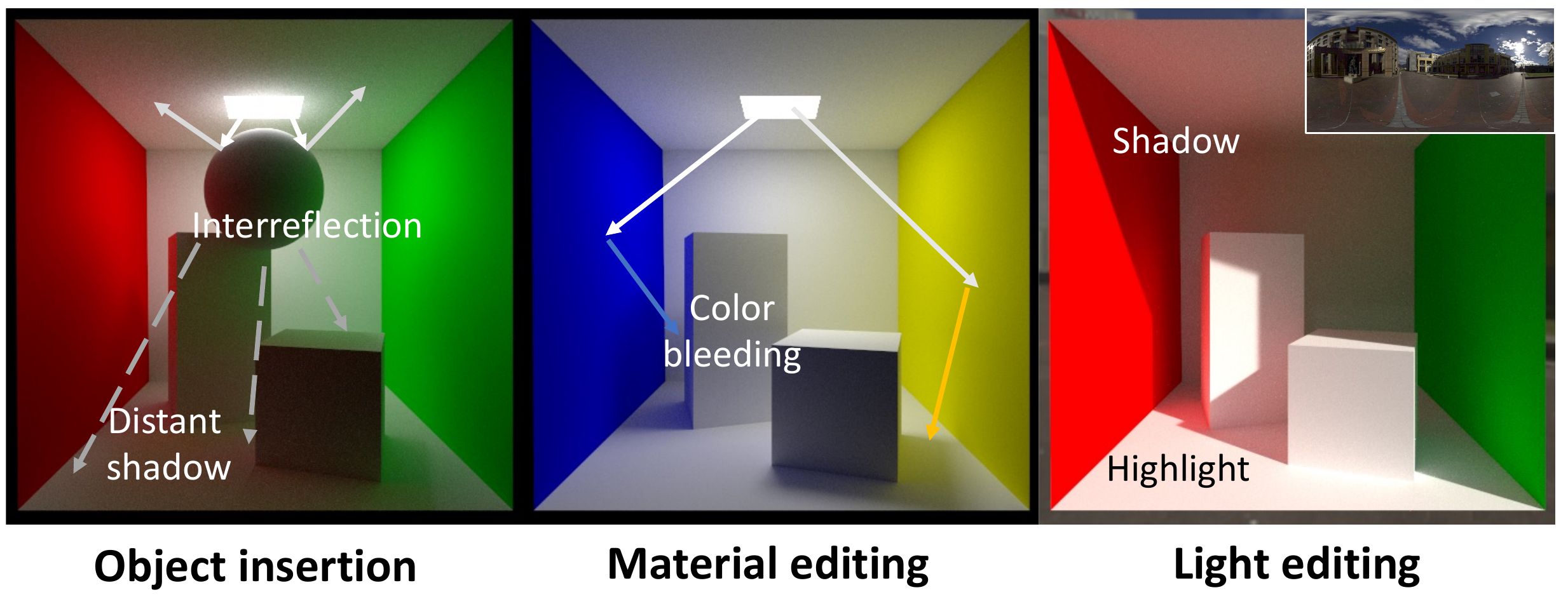}
\end{minipage}\hfill
\begin{minipage}[c]{0.38\linewidth}
\caption{
\mc{Image editing must explicitly predict light sources to account for global effects such as distant shadows and interreflections due to inserted objects, color bleeding on far surfaces due to edited materials and light shafts by opening a window.} }
\label{fig:illustration}
\end{minipage}
\vspace{-0.3cm}
\end{figure}

Several recent works estimate {\em lighting} in indoor scenes \cite{li2020inverse,srinivasan2020lighthouse,gardner2019parametric,wang2021learning}, but achieving the above outcomes requires estimating and editing {\em light sources}. While both are highly ill-posed for single-image inputs, we posit that the latter presents fundamentally different and harder challenges for computer vision. First, it requires disentangling the individual contributions of both visible and invisible light sources, independent of the effects of geometry and material. Second, it requires reasoning about long-range effects such as interreflections, shadows and highlights, while also being precise about highly localized 3D shapes, spectra, directions and bandwidths of light sources, where minor errors can lead to global artifacts due to the above distant interactions. Third, it requires photorealistic re-rendering of the scene despite only partial observations of geometry and material, while handling complex light transport. \mc{Figure \ref{fig:illustration} illustrates a few such challenges.}

We solve the above challenges by bringing together a rich set of insights across physically-based vision and neural rendering. Given a single LDR image of an indoor scene, with predicted depth map and masks for visible lights, we propose an inverse rendering method to estimate a {\em parametric model} of both visible and invisible light sources (in addition to a per-pixel SVBRDF). Beyond a 3D localization of light sources, our modeling accurately supports their physical properties such as geometry, color, directionality and fall-off. Next, we design a {\em neural differentiable renderer} that judiciously uses classical methods and learned priors to synthesize high-quality images from predicted reflectance and light sources. We accurately model long-range light transport through a physically-based Monte Carlo ray tracer with a learned shadow denoiser to render direct irradiance and visibility, which combines with an indirect irradiance network to predict local incoming lighting at every pixel. Our neural renderer injects the inductive bias of physical image formation in training, while allowing rendering and editing of global light transport from partial observations, as well as optimization to refine editing outputs.

Our parametric light source estimation and physically-based neural renderer allow \mc{intuitive editing of multiple lamps and windows, with their global effects handled explicitly in scene relighting.} In Fig.~\ref{fig:teaser} (c,d), we turn off individual visible and invisible lamps. Beyond standard object insertion of prior works (e), we visualize inserted objects by ``turning on’’ a new lamp (g,h) or ``opening’’ a window with incoming sunlight (i, j). In each case, global effects such as highlights, shadows and interreflections are accurately created for the entire scene by the neural renderer, and are also properly handled when we edit material properties of scene surfaces (f). \zl{In the accompanying \href{https://drive.google.com/file/d/1IuMuJ4QyVGIWNhN_HhwOhyJuQ7Ud26b5/view?usp=sharing}{video}, we show that these editing effects are consistent as we move virtual objects and light sources, or gradually change materials.} These abilities significantly surpass those of prior methods for intrinsic decomposition or inverse rendering. \mc{As summarized in Tab.~\ref{tab:editApps} and elaborated in Sec.~\ref{sec:relatedworks}, the proposed method is the first to allow a broad range of single image scene relighting abilities in the form of inserting objects, changing complex materials and editing light sources, with consistent global interactions.}

\begin{table}[!!t]
    \centering
\begin{minipage}[c]{0.6\linewidth}
\includegraphics[width=\columnwidth]{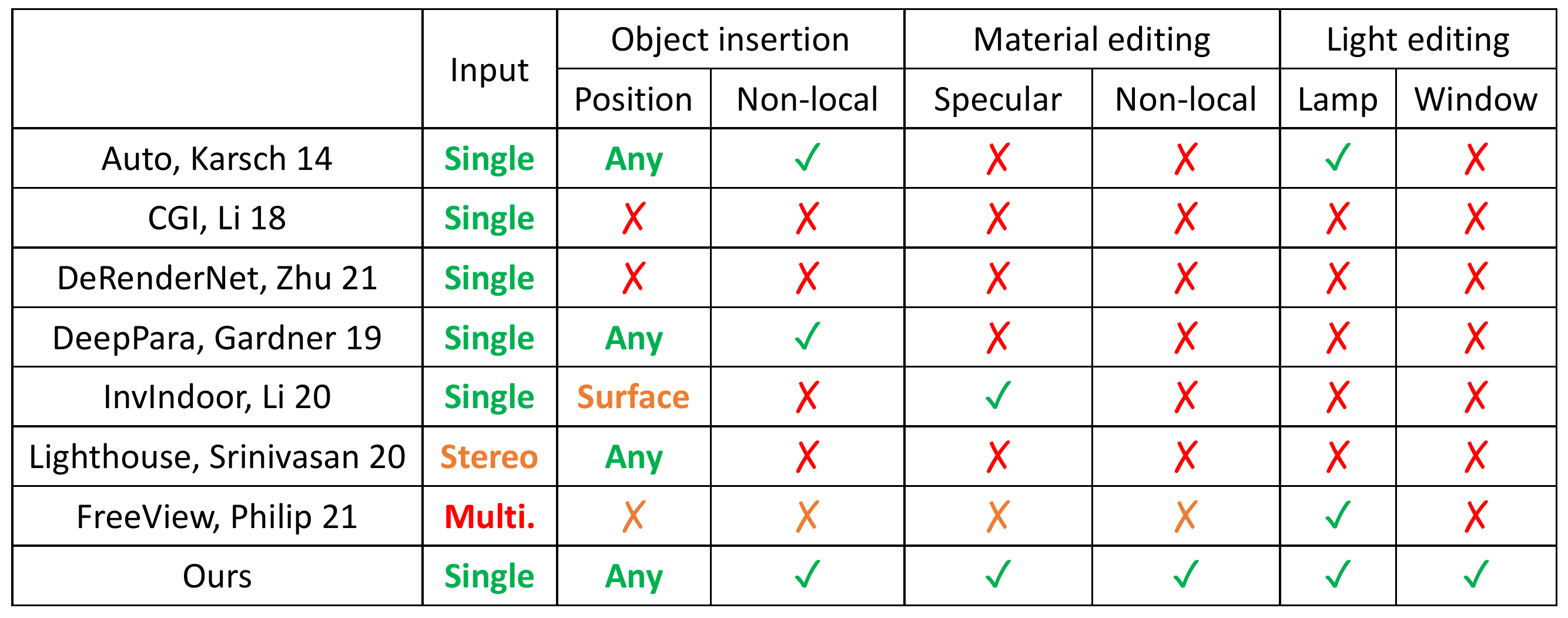}
\end{minipage}\hfill
\begin{minipage}{0.38\linewidth}
\caption{Compared to prior works on inverse rendering, ours enables full scene relighting with global effects for inserted objects, edited materials or light sources. Also see Figs.~\ref{fig:teaser} and \ref{fig:illustration}. }
\label{tab:editApps}
\end{minipage}
\vspace{-0.5cm}
\end{table}

\vspace{-0.2cm}
\section{Related Work}
\label{sec:relatedworks}
\vspace{-0.2cm}

\paragraph{Inverse rendering.} 
Inverse rendering seeks to estimate factors of image formation (shape, materials and lighting) \cite{Marschner1998InverseRF}, which has traditionally required multiple images and controlled setups \cite{goldman2010shape,chandraker2014shapematerial,xia2016recovering,bi2020deep}. Several single-image works on material acquisition \cite{li2017selfaugmented,li2018material}, or object-level shape and reflectance reconstruction use known \cite{oxholm2012shape,johnson2011natural} or semi-controlled lighting \cite{li2018learning}. We consider a complex indoor scene under unknown illumination and jointly estimate its geometry, material and lighting from a single LDR image. Intrinsic decomposition \cite{barrow1978intrinsic,shen2011intrinsic,bell2014intrinsic,bi20151,li2018cgintrinsics,li2018learningintrinsic} decomposes an image into Lambertian reflectance and diffuse shading. \zl{ Recent works also seek to predict a shadow map \cite{zhu2021derendernet} or separate global and indirect illumination \cite{meka2021real}}. Several deep learning methods estimate complex SVBRDFs and lighting \cite{sengupta2019neural,li2020inverse}. But none of the above can estimate or edit light sources. We instead propose a novel physically-based 3D light source representation and neural rendering framework that estimates and edits individual light sources \zl{with distant shadows and global illumination being explicitly handled.}

\begin{figure*}[!!t]
\centering
\includegraphics[width=\textwidth]{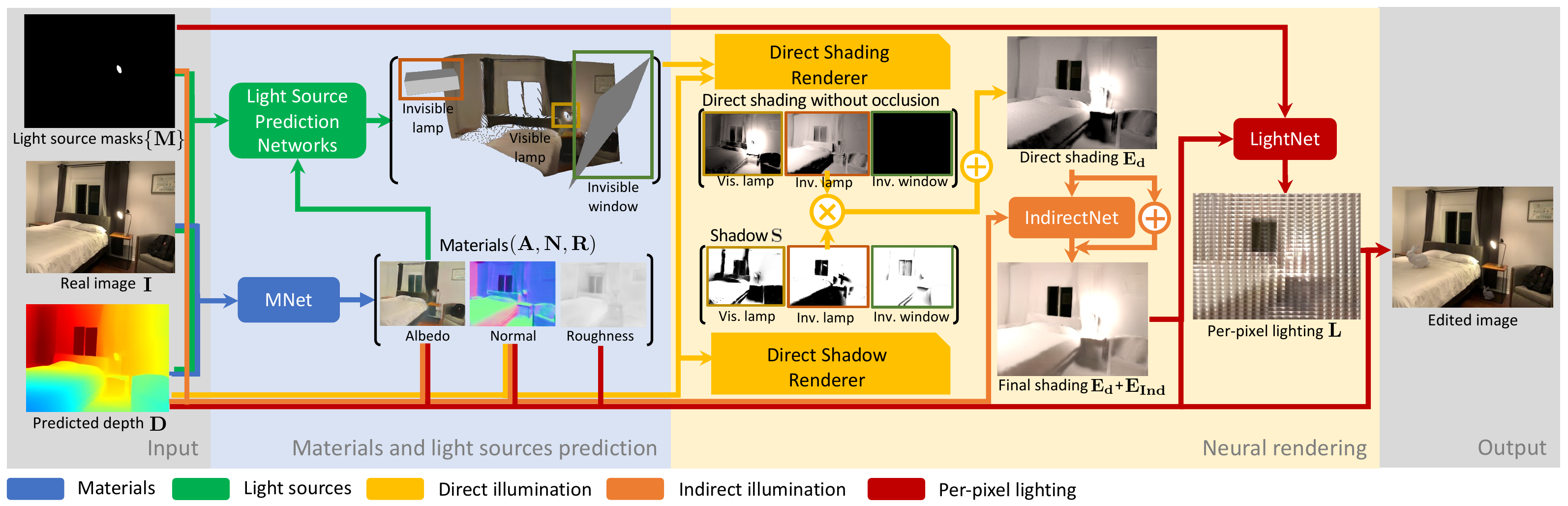}
\vspace{-0.8cm}
\caption{Overview of our method. We start from an RGB image. The depth map and visible light source masks can be estimated from the RGB image or given as additional inputs. We estimate per-pixel material parameters (albedo, normal, roughness) using a network (blue). Next, we estimate light sources (windows and lamps, visible and invisible) using four networks (green). At this point, we can edit the scene representation (lights, materials, depth). To render the edited representation back into an image, we use a neural renderer consisting of three modules: direct shading, shadow (yellow), and indirect shading module (orange). The result is per-pixel shading (diffuse irradiance), which can be turned into per-pixel lighting (a grid of incoming radiance environment maps) using another network (red).}
\label{fig:pipeline}
\vspace{-0.6cm}
\end{figure*}

\vspace{-0.3cm}
\paragraph{Lighting estimation and representation.}
Many single image approaches estimate lighting globally as an environment map \cite{debevec1998rendering,gardner2017indoor,legendre2019deeplight}, which cannot express the complex spatial variation of indoor illumination. Some recent works model spatial variations as per-pixel environment maps \cite{barron2015sirfs,zhou2019glosh,garon2019fast,li2020inverse}, or volumes \cite{srinivasan2020lighthouse,wang2021learning}. However, these are non-parametric  representations, which can mainly be used for object insertion, while we estimate editable windows and lamps (visible and invisible) with physically meaningful properties (such as position, direction, shape, size and intensity).
Gardner et~al. \cite{gardner2019parametric} predict a fixed number of spherical Gaussian lobes to approximate indoor light sources but do not handle light editing or its global effects. Zhang et al. recover the geometry and radiance of an empty room but cannot handle furniture inside \cite{zhang2016emptying}. Karsch et al. reconstruct geometry, reflectance and lighting but do not model windows and invisible scene contributions, require extensive user inputs \cite{karsch-siga-11} or face artifacts from imperfect heuristics or optimization \cite{karsch-tog-14}. In contrast, our physically-based neural renderer synthesizes photorealistic images with complex light transport, to enable relighting, light source insertion and removal from a single image.

\vspace{-0.3cm}
\paragraph{Neural rendering and relighting.}
NeRF \cite{mildenhall2020nerf} and other volumetric neural rendering approaches have achieved photo-realistic outputs, but usually limited to view synthesis \cite{mildenhall2020nerf,yu2020pixelnerf,liu2020neural}. A few recent works \cite{bi2020deepreflectance,bi2020neural,boss2020nerd,srinivasan2020nerv,xiang2021neutex}  handle relighting, but use a per-object optimization from a large set of images. Philip et~al. \cite{philip2019multi} demonstrate relighting for outdoor scenes but require multiple images. Concurrent to our work, Philip et~al. \cite{philip2021free} consider indoor relighting, but require a large number of high-resolution RAW images, cannot reconstruct complex directional sunlight and do not support material editing and object insertion with their neural renderer. As shown in Fig.~\ref{fig:illustration} and Tab.~\ref{tab:editApps}, our modeling and neural rendering enable applications not possible for prior works, such as light source insertion and removal, or insertion of virtual objects and changing of materials with non-local effects, with a single LDR image.
\vspace{-0.2cm}
\section{Material and Light Source Prediction}
\label{sec:mglPrediction}
\vspace{-0.2cm}

Our overall framework is summarized in Fig.~\ref{fig:pipeline}. In this section, we describe our novel, physically meaningful and editable representations, while Sec.~\ref{sec:envRenderLayer} describes our neural renderer that is differentiable with respect to light sources to facilitate training and editing of complex light transport.

\vspace{-0.3cm}
\paragraph{Per-pixel normal and material prediction}
We first train a U-net similar to \cite{li2020inverse} to predict material parameters per pixel of the input image: diffuse albedo ($\mathbf{A}$), normal ($\mathbf{N}$) and roughness ($\mathbf{R}$), following the SVBRDF model of \cite{karis2013unreal}. The input to the network is a $240 \times 320$ LDR image ($\mathbf{I}$) and its corresponding depth map ($\mathbf{D}$), which in our case can be predicted by a state-of-the-art monocular depth prediction network \cite{ranftl2021vision}.  We predict the normals directly, instead of computing them as the normalized gradient of depth to avoid artifacts and discontinuities. Thus, our prediction is given by $\{ \mathbf{A}, \mathbf{{N}}, \mathbf{R} \} = \mathbf{MNet(I, D)}$.

\vspace{-0.2cm}
\subsection{Light Source Representation}
\label{sec:lightrepresentation}
\vspace{-0.1cm}

To enable indoor scene relighting from a single LDR image, we need light source representations that are editable, expressive enough for different types of lighting and realistic enough for convincing rendering of complex scenes. We model the radiance and geometry of two types of common indoor light sources with very different properties: (a) {\em windows} that usually cover large areas and may induce strong directional lighting from the sun, and (b) {\em lamps} that tend to be small but with more complex geometry.

\begin{figure}[t]
\centering
\begin{minipage}[c]{0.6\linewidth}
\includegraphics[width = \columnwidth]{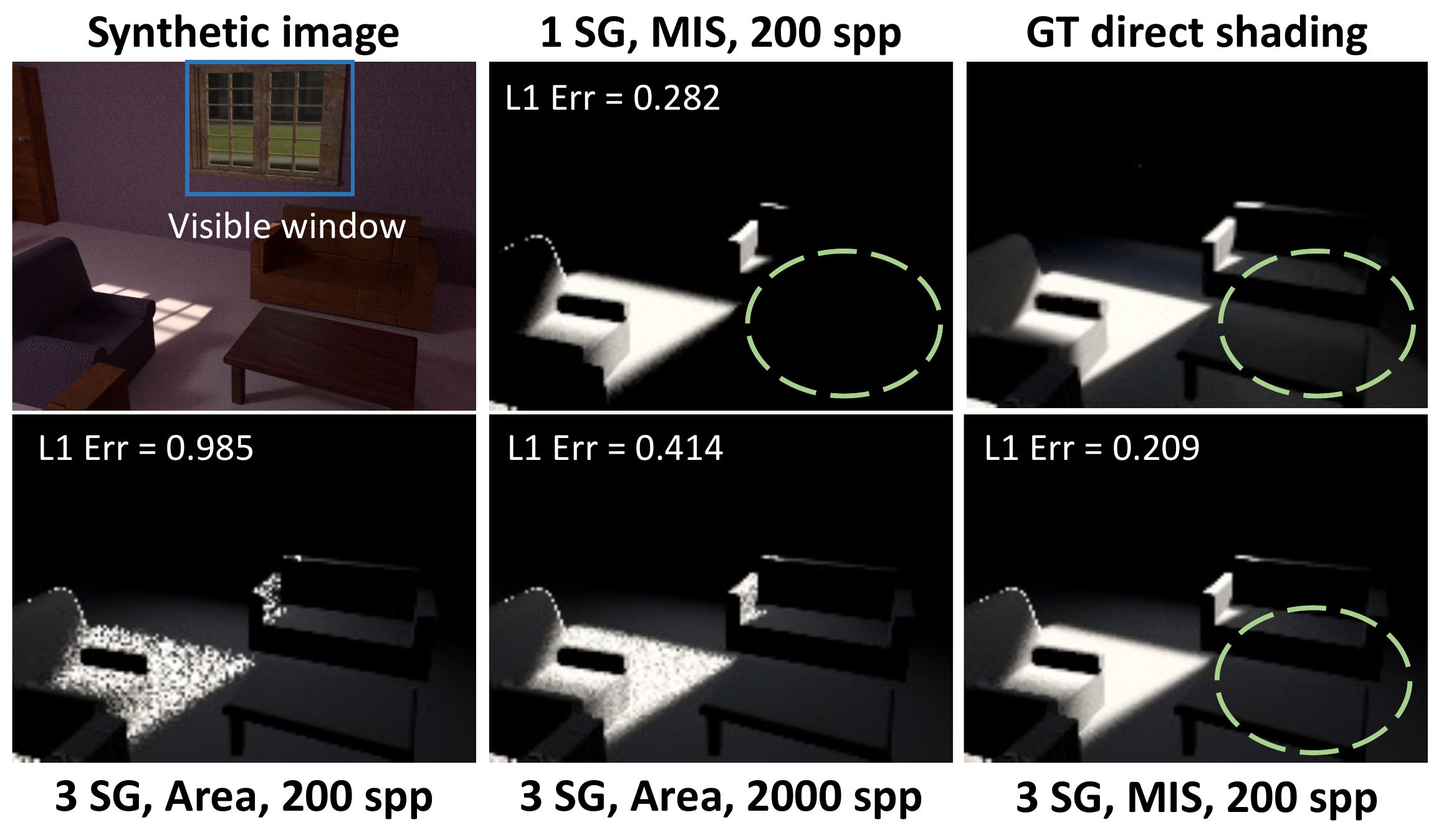}
\end{minipage}\hfill
\begin{minipage}[c]{0.38\linewidth}
\vspace{-0.4cm}
\caption{Comparisons of direct shading rendered from different window representations with different sampling methods. We show that our 3 SGs models ambient lighting much better than a single SG, as shown in the green circle, and MIS sampling leads to much less noise compared to sampling window area uniformly.}
\label{fig:winRep}
\end{minipage}
\vspace{-0.3cm}
\end{figure}

\vspace{-0.3cm}
\paragraph{Radiance.}
The emitted radiance of lamps can be modeled by a standard Lambertian model, where every surface point with intensity $\mathbf{w}$ emits light uniformly into its hemisphere. However, the radiance distribution of windows can be strongly directional due to sunlight coming through on a clear day, which is important for capturing realistic indoor lighting but often neglected by prior methods \cite{straub2019replica,philip2021free,srinivasan2020lighthouse}. 
A recent work \cite{wang2021learning} models directional lighting with a single spherical Gaussian (SG), but as shown in Fig.~\ref{fig:winRep}, cannot recover ambient effects leading to suboptimal rendering. Instead, we model the directional distribution of window radiance with 3 SGs corresponding to the sun, sky and ground. Each SG is defined by three parameters $\mathcal{G}_{\mathbf{k}}=(\mathbf{w_k}, \lambda_{\mathbf{k}}, \mathbf{d_k})$, for intensity, bandwidth and direction of lighting. For a ray in direction $\mathbf{l}$ that hits the window, its intensity is $\mathbf{L}_\mathcal{W}(\mathbf{{l}}) = \sum_{\mathbf{k}} \mathbf{w_k}\exp\big(\lambda_{\mathbf{k}}(\mathbf{d_k}\cdot\mathbf{l}-1)  \big)$, where $k \in \{\text{sun, sky, grnd}\}$.
 Fig.~\ref{fig:winRep} shows that our representation with multiple importance sampling leads to direct shading close to the ground-truth.

\begin{figure}[t]
\centering
\begin{minipage}[c]{0.6\linewidth}
\includegraphics[width = \columnwidth]{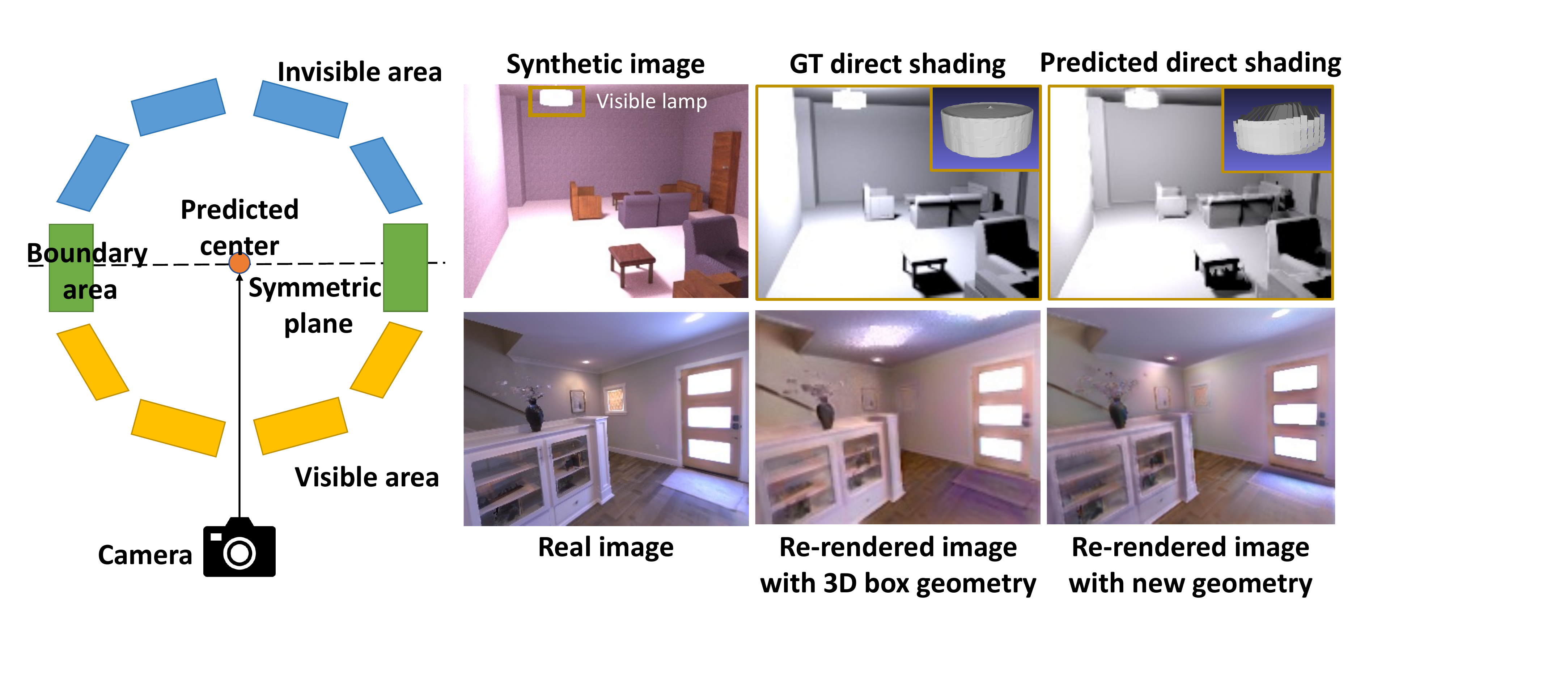}
\end{minipage}\hfill
\begin{minipage}[c]{0.38\linewidth}
\vspace{-0.4cm}
\caption{A demonstration of our visible lamp geometry representation. Our representation for visible lamps is much less likely to cause highlight artifacts and wrong shadows compared to a standard 3D bounding box.}
\label{fig:lampRep}
\end{minipage}
\vspace{-0.4cm}
\end{figure}

\vspace{-0.3cm}
\paragraph{Geometry.}
Window geometry can be simply approximated by a rectangle $\{\mathbf{c, x, y}\}$, where $\mathbf{c}$ is the center and $\mathbf{x, y}$ are the two axes. However, indoor lamps present more diverse geometry. Naively representing a lamp with a 3D bounding box $\{\mathbf{c, x, y, z}\}$ works for invisible lamps in the scene, but it often leads to artifacts for visible lamps, as the imperfect shape generates incorrect highlights. Therefore, we carefully design a new visible lamp representation as shown in Fig. \ref{fig:lampRep}. We first identify the visible surface based on the depth $\mathbf{D}$ and lamp segmentation mask $\mathbf{M}_{\mathcal{L}}$, reconstruct the invisible surface by reflecting the visible surface with respect to the lamp center $\mathbf{c}$ and then add the boundary area. As shown in Fig. \ref{fig:lampRep}, our new representation can effectively constrain the lamp geometry and achieve realistic rendering without highlight artifacts for difficult real world examples. More details are in the supplementary.

\vspace{-0.2cm}
\subsection{Light Source Prediction}
\label{sec:lightprediction}
\vspace{-0.1cm}

We use four neural networks to predict visible and invisible light sources for the lamp and window categories. For visible light sources, the inputs to the network include extra instance segmentation masks for visible lamps and windows that are turned on in the scene. We can obtain the instance segmentation mask by either fine-tuning a Mask R-CNN\cite{he2017mask} for our dataset, combined with a graph-cut based post processing to refine the boundaries, or manually draw the masks. While this is not our main focus, we include both qualitative and quantitative analysis in the supplementary. Let $\mathbf{M}_\mathcal{W}$ be a mask for a window and $\mathbf{M}_\mathcal{L}$ be a mask for a lamp. For each visible window and lamp, we have
\begin{eqnarray*}
\{\mathbf{c, w} \}  &=& \mathbf{VisLampNet(I, A, D, M_\mathcal{L} ) }, \label{eq:visLampNet} \\
\{\mathbf{c, x, y}, \mathcal{G}_{\text{sun}}, \mathcal{G}_{\text{sky}}, \mathcal{G}_{\text{grnd}} \} &=&  \mathbf{VisWinNet(I, A, D, M_\mathcal{W} ) }. \label{eq:visWinNet}
\end{eqnarray*}

\begin{figure}[t]
\centering
\begin{minipage}[c]{0.6\linewidth}
\includegraphics[width=\columnwidth]{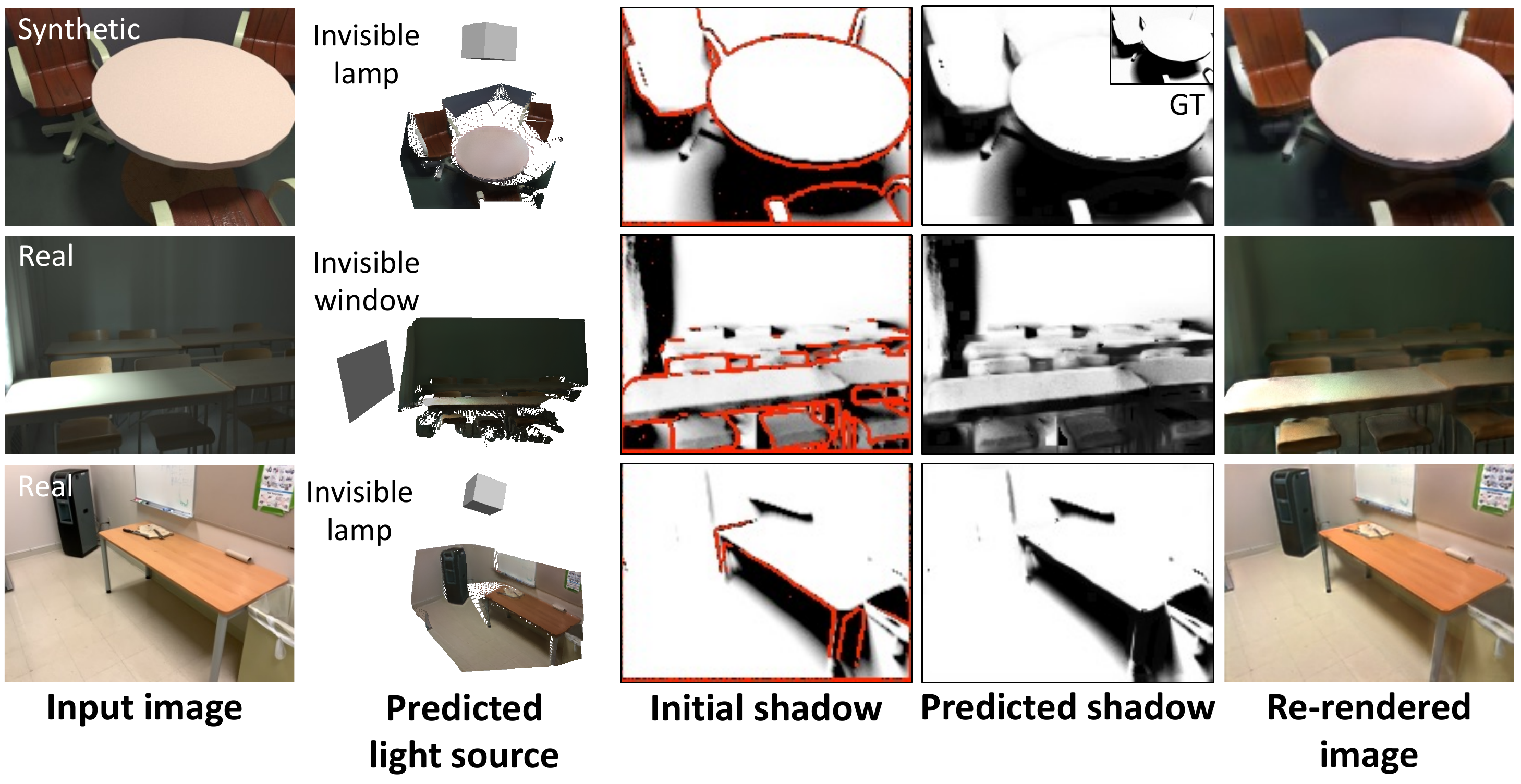}
\end{minipage}\hfill
\begin{minipage}[c]{0.38\linewidth}
\vspace{-0.4cm}
\caption{Direct rendering shadows with ray tracing leads to boundary artifacts as shown in the red color in the third column. In contrast, our trained depth-based shadow renderer achieves high-quality shadows for both real and synthetic scenes, leading to re-rendered images that closely match the inputs.}
\label{fig:shadow}
\end{minipage}
\vspace{-0.4cm}
\end{figure}

We assume one invisible lamp as a 3D bounding box and one invisible window. These are deliberate simplifications: while invisible lights can contribute significant illumination, they are hard to infer using only indirect cues. We limit the expressivity of the representation to account for this ill-posedness and find it to be a good choice in practice\footnote{The real scene in Fig.~\ref{fig:teaser} has 4 invisible lamps and the last real scene in Fig.~\ref{fig:shadow} has 2. In both cases, we achieve reasonable approximation with one invisible lamp.}. When a scene has no invisible light sources, their predicted intensity is close to zero, as shown in Fig.~\ref{fig:pipeline} and Fig.~\ref{fig:optim}. To learn a better separation of the contributions of visible and invisible light sources, we provide a mask $\mathbf{M = \sum_{\mathcal{W}}M_\mathcal{W} + \sum_{\mathcal{L} } M_\mathcal{L} }$ of all visible sources to the invisible light estimation networks:
\begin{eqnarray*}
\{\mathbf{c, x, y, z}\} &=& \mathbf{InvLampNet(I, A, D, M) }, \\
\{\mathbf{c, x, y}, \mathcal{G}_{\text{sun}}, \mathcal{G}_{\text{sky}}, \mathcal{G}_{\text{grnd}} \} &=& \mathbf{InvWinNet(I, A, D, M) }.
\end{eqnarray*}

\section{Neural Rendering Framework}
\label{sec:envRenderLayer}
\vspace{-0.2cm}

In order to achieve photorealistic editing of indoor lighting, we need a rendering framework that can handle complex light transport typical for indoor scenes, including sharp directional lighting, hard and soft shadows, global illumination and specular materials. While existing differentiable path tracers can handle all these effects, they are computationally expensive. Even more importantly, they require the full reconstruction of material, geometry and lighting of the entire indoor scene, including its invisible parts. 

To address these limitations, we introduce a neural rendering framework that combines the advantages of physically-based rendering and learning-based rendering, which works with our novel light source representations, does not require reconstruction of invisible scene surfaces, achieves high performance, and supports differentiability. Our framework, illustrated in Fig.~\ref{fig:pipeline} (right), has 3 modules: (1) a physically-based direct shading module that computes the direct irradiance at a surface point from each light source through Monte Carlo sampling; (2) a hybrid shadow module by casting rays against a mesh constructed from depth information and . 
(3) an indirect shading module that predicts indirect shading (global illumination) from direct shading. (4) a per-pixel lighting reconstruction module that turns the shading, materials and geometry predictions into per-pixel environment map, which can be used to insert specular objects.

Our direct shading and shadows are computed based on standard techniques, while global illumination and per-pixel lighting are predicted by networks. The reason is that in the absence of full scene reconstruction (i.e. invisible parts), global illumination can only be computed heuristically, which is suited for neural networks. Conversely, direct illumination and non-local shadowing can be efficiently computed by standard techniques, but remain tricky for neural methods.

\vspace{-0.2cm}
\subsection{Direct Shading Rendering Module}
\label{sec:directShading}
\vspace{-0.1cm}

We use inspiration from physically-based rendering \cite{pharr2016physically} to sample the surface of each light source and connect those samples to the scene points. Formally, let $\mathbf{p}$ be a shading point and $\mathbf{q}$ be a point uniformly sampled on the light surface, with $\mathbf{p\!\!\rightarrow\!\!q}$ the unit vector from $\mathbf{p}$ to $\mathbf{q}$. The direct shading $\mathbf{E_j}$ caused by light source $\mathbf{j}$ is computed as:
\small
\vspace{-0.2cm}
\begin{equation}
\mathbf{E_j} (\mathbf{p})=
\frac{\mbox{area}(\mathbf{j})}{N_\mathbf{j}}
\sum_{\mathbf{q}} \frac{\mathbf{L_j}(\mathbf{q\!\!\rightarrow\!\!p})\max(\cos\theta_{\mathbf{p}}\cos\theta_{\mathbf{q}}, 0)}{|| \mathbf{q}-\mathbf{p}||_2^2},\label{eq:areaSample}  
\vspace{-0.1cm}
\end{equation}
\normalsize
where $\cos\theta_{\mathbf{p}} = \mathbf{p\!\!\rightarrow\!\!q}\cdot\mathbf{N(p)}$, $\cos\theta_{\mathbf{q}} = \mathbf{q\!\!\rightarrow\!\!p}\cdot\mathbf{N(q)}$ and $N_{\mathbf{j}}$ is the number of samples for light source $\mathbf{j}$. While our Monte Carlo estimation in \eqref{eq:areaSample} converges fast for lamps, it is not optimal for high-frequency directional sunlight coming through windows, since only when $\mathbf{q\!\!\rightarrow\!\! p}$ aligns with the sun direction, will the $\mathbf{L}(\mathbf{q\!\!\rightarrow\!\! p})$ return a significant contribution. To tackle this issue, with $\mathbf{Pr}(\mathbf{l})$ the probability of sampling direction $\mathbf{l}$ from $\mathcal{G}_{sun}$, we also generate samples according to the angular distribution of $\mathcal{G}_{\text{sun}}$:
\small
\vspace{-0.2cm}
\begin{equation}
\mathbf{E_j} (\mathbf{p})=
\sum_{\mathbf{l}} \frac{\mathbf{L_j}(\mathbf{l})\mathbf{I_j(l)}\max(\cos\theta_{\mathbf{p}}, 0)}{N_{\mathbf{j}} \mathbf{Pr(l)} },\label{eq:angSample}  
\vspace{-0.1cm}
\end{equation}
\normalsize
where $\mathbf{I_j(l)}$ is an indicator function to detect if ray $\mathbf{l}$ starting from $\mathbf{p}$ can hit the window plane. Note that both \eqref{eq:areaSample} and \eqref{eq:angSample} are unbiased but with different variances, which we combine with multiple importance sampling (MIS) \cite{veach1997robust}. Details are in supplementary. Fig.~\ref{fig:winRep} compares the direct shading of a window, where we observe that our MIS method can render high-quality direct shading with much fewer samples, \zl{which makes training with rendering loss possible}.

\begin{table}[!!t]
\footnotesize
\centering
\begin{minipage}[c]{0.5\linewidth}
\centering
\begin{tabular}{|c|c|c|}
\hline 
& Ray traced  & Ours  \\ 
\hline
$L_2$ & 0.011  & 0.005 \\
\hline
\end{tabular}                 
\end{minipage}\hfill
\begin{minipage}[c]{0.5\linewidth}
\caption{
Shadow rendering error with or w/o network inpainting. 
\label{tab:shadow}}
\end{minipage}
\vspace{-0.4cm}
\end{table}

\vspace{-0.2cm}
\subsection{Depth-based Shadow Rendering Module}
\label{sec:shadow}
\vspace{-0.1cm}

Recall that in the above shading computation, $\mathbf{E_j}, j\in \{\mathcal{W}\}\cup\{\mathcal{L}\}$ does not consider visibility and therefore cannot handle shadows. We could check visibility by ray-tracing during the Monte Carlo sampling above, but this causes artifacts due to incomplete geometry, as shown in Fig. \ref{fig:shadow}. We instead design a depth-based shadow rendering framework that combines Monte Carlo ray tracing with deep network inpainting and denoising. Note that our shadow modules are not differentiable, as this is not necessary for our application: we train our network on a synthetic dataset, where it is provided with the ground-truth supervision of direct shading without the shadow effects, so back-propagation of error through the shadow renderer is not used during training.

Our approach first creates a mesh from the depth map, and then uses a GPU-based ray-tracer to cast shadow rays from surfaces to light sources. To address the boundary artifacts, we first modify the renderer to detect the occlusion boundaries, then train a CNN to fill in the shadow at these regions. This hybrid approach outperforms both pure ray-tracing and a CNN trained to clean up the entire ray-traced shadow image. Formally, let $\mathbf{S^{Init}}$ be the initial shadow image rendered from depth map $\mathbf{D}$ and let $\mathbf{M^{S}}$ be the mask for occlusion boundaries. 
\begin{equation}
\small
\mathbf{S} = \mathbf{M^{S} \cdot DShdNet(S^{Init}, D, N) } 
+ \mathbf{(1 - M^{S})\cdot S^{Init}}.
\end{equation}
The total direct shading from all sources is $\mathbf{E_d} = \sum_{\mathbf{j}} \mathbf{E_j S_j}$. 
As seen in Fig.~\ref{fig:teaser},~\ref{fig:shadow} and~\ref{fig:indirect}, our framework can render higher quality soft and hard shadows that are closer to the ground-truths compared to a standard ray tracer. Tab.~\ref{tab:shadow} shows that our CNN reduces the shadow error by more than $50\%$. 
 
\vspace{-0.2cm}
\subsection{Indirect Shading Prediction}
\label{sec:indirect}
\vspace{-0.1cm}

\begin{figure}[t]
\centering
\begin{minipage}[c]{0.58\linewidth}
\includegraphics[width=\columnwidth]{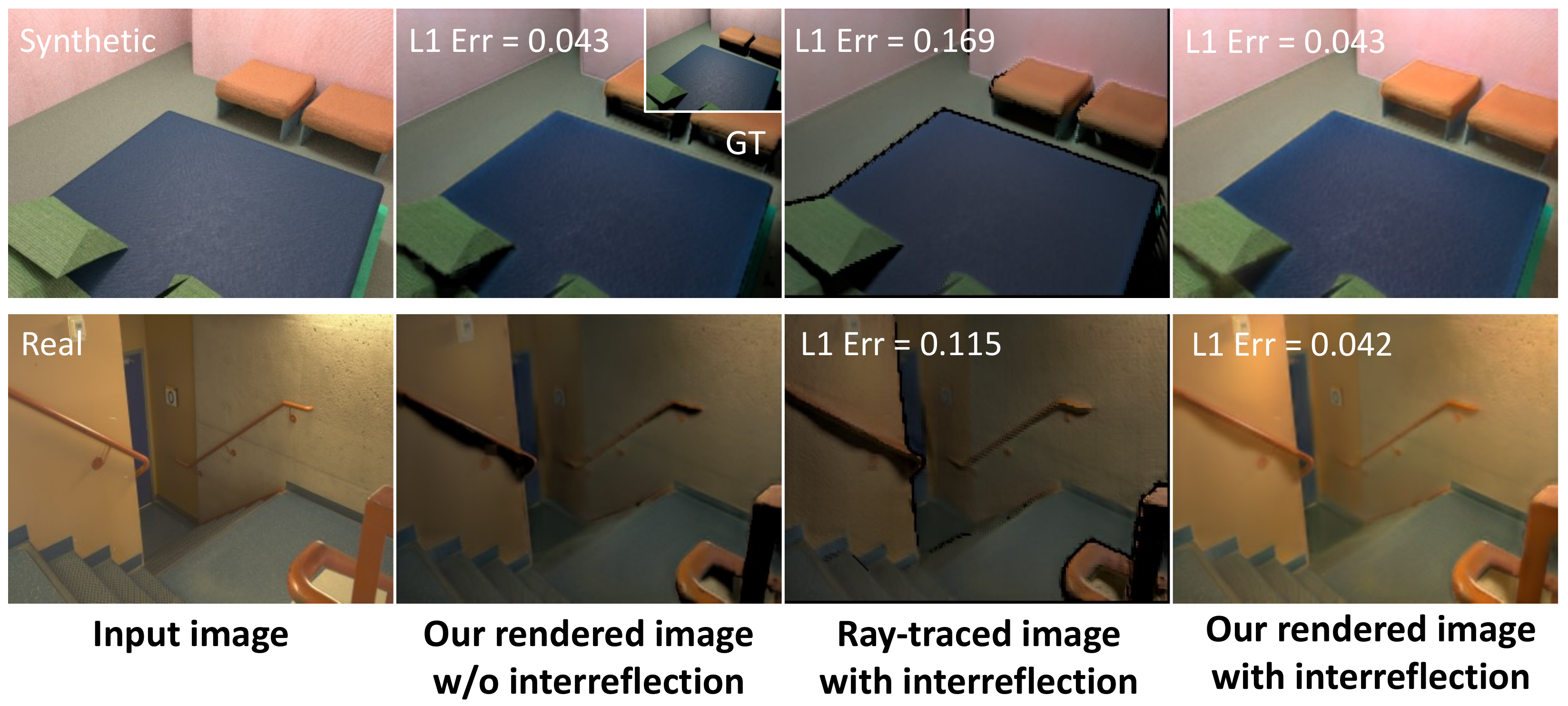}
\end{minipage}\hfill
\begin{minipage}[c]{0.40\linewidth}
\vspace{-0.4cm}
\caption{\mc{Our neural renderer models both direct and indirect illumination accurately, while a path tracer using single-view reconstruction cannot model indirect illumination with incomplete geometry and materials and has artifacts near occlusion boundaries.} }
\label{fig:indirect}
\end{minipage}
\vspace{-0.4cm}
\end{figure}

To render indirect shading with a standard physically-based renderer, we would need to reconstruct invisible geometry and materials, which is challenging. Instead, we train a 2D CNN to predict indirect shading in screen space. A similar idea has been adopted by recent work \cite{xin2020lightweight}. We use a network with large receptive field covering the entire image to model non-local inter-reflections. Our indirect shading is $\mathbf{E_{Ind}} = \mathbf{IndirectNet(E_{d}, D, N, A)}$, which is added to the direct shading for the final shading prediction. In Fig.~\ref{fig:indirect}, we compare the indirect illumination rendered by our network and by a standard path tracer by first building a mesh from the depth map and then texturing the mesh with predicted materials. Quantitative and qualitative results on real and synthetic examples show that our neural rendering layer renders both direct and indirect illumination accurately, while a path tracer cannot handle indirect illumination with partial geometry, leading to an image with similar intensity as one with direct illumination only.   

\vspace{-0.2cm}
\subsection{Predicting Lighting From Shading}
\label{sec:localLighting}
\vspace{-0.1cm}

The above framework cannot yet handle specular reflectance, which motivates us to add another network to infer spatially and directionally varying incoming lighting $\mathbf{L}$, taking the above shading (irradiance) $\mathbf{E}$ as input. We follow \cite{li2020inverse} to predict a grid of environment maps. We use a similar network architecture but replace the input image $\mathbf{I}$ with the shading $\mathbf{E}$ so that the predicted \emph{local} lighting is a function of our lighting representation: $\mathbf{L} = \mathbf{LightNet(E, M, A, N, R, D)}$. The resulting incoming radiance field $\mathbf{L}$ can be used to render specular materials, as shown in Fig.~\ref{fig:garon}  and Fig.~\ref{fig:garonPlus} in Sec.~\ref{sec:experiment}.

\vspace{-0.2cm}
\section{Implementation Details}
\label{sec:implementation}
\vspace{-0.2cm}

\begin{figure}[t]
\centering
\begin{minipage}[c]{0.6\linewidth}
\includegraphics[width=\columnwidth]{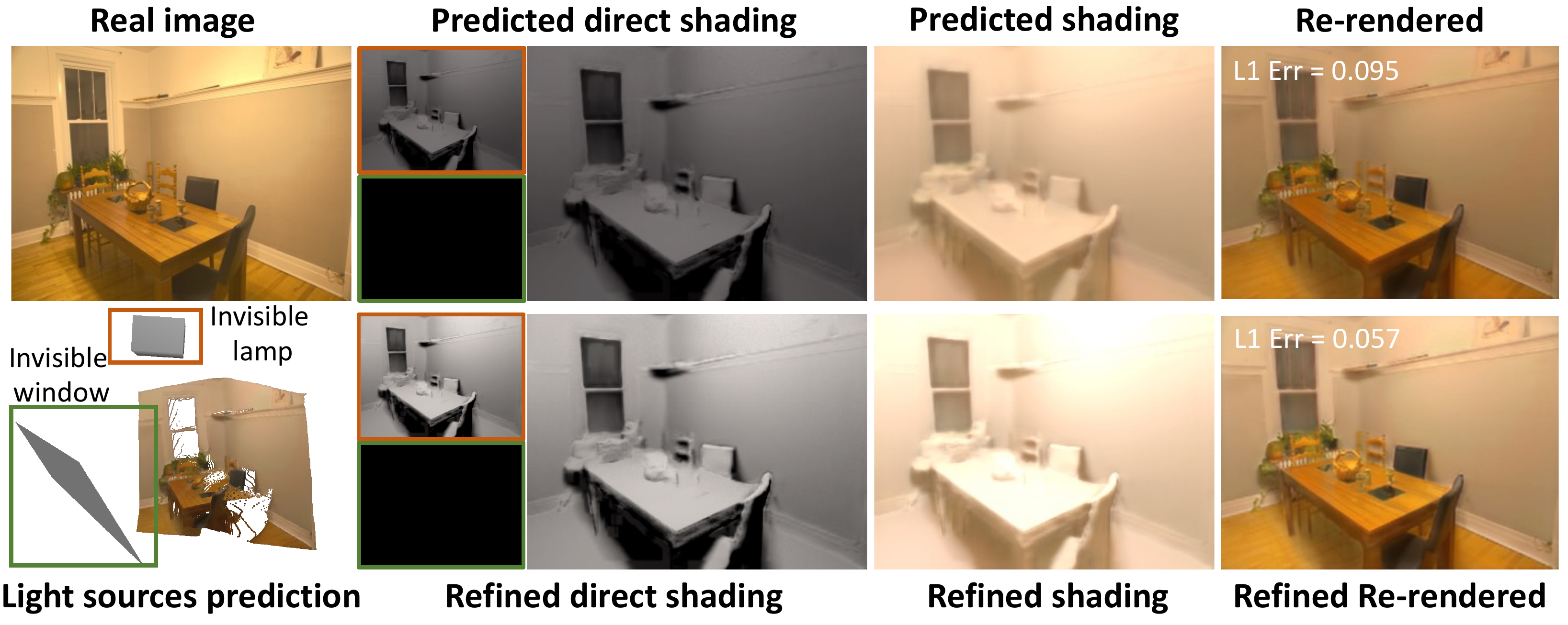}
\end{minipage}\hfill
\begin{minipage}[c]{0.38\linewidth}
\caption{Comparisons of light source prediction and rendering before and after the optimization on a real scene. Our neural renderer allows using the rendering loss to learn and refine light source intensity and direction}
\label{fig:optim}
\end{minipage}
\vspace{-0.4cm}
\end{figure}

\paragraph{Dataset}
\mc{We train on OpenRooms \cite{li2020openrooms} -- a large-scale synthetic indoor dataset for inverse rendering -- which is unique among currently available datasets in providing ground truths for all our outputs, such as light source geometry, per-light source shadings (with and without occlusion) and per-light source shadows. Thus, it allows to train each module separately, significantly simplifying training.}

\vspace{-0.3cm}
\paragraph{Optimized light source parameters} We augment the OpenRooms dataset with optimized light source parameters $\{\mathcal{G}_{\text{sun}}, \mathcal{G}_{\text{sky}}, \mathcal{G}_{\text{grd}}\}$ for windows, leading to sharper and more interpretable predictions. To compute those, we minimize the $L_1$ difference between the rendered direct shading without occlusion $\mathbf{E_j}$, $j\in\{\mathcal{W}\}$ and its corresponding ground truth, through our differentiable Monte Carlo rendering module (Sec \ref{sec:directShading}). Further details are in supplementary. The optimized direct shading is seen in Fig.~\ref{fig:winRep} to closely match the ground truth. 

\vspace{-0.3cm}
\paragraph{Losses}
To train $\mathbf{MNet}$, we use $L_2$ loss on the albedo, normal and roughness. The loss function for light source prediction is the sum of a rendering loss ($\mathbf{Loss_{ren}}$), a geometry loss ($\mathbf{Loss_{geo}}$), and a light source loss ($\mathbf{Loss_{src}}$). For  $\mathbf{Loss_{ren}}$, we define it to be the $L_1$ distance between the rendered direct shading $\mathbf{E_j}$ and its ground-truth direct shading, both without shadows applied. For $\mathbf{Loss_{geo}}$, we uniformly sample sets of points $\{\mathbf{\mathbf{q}}\}$ from the ground-truth and predicted light source geometry to compute their RMSE Chamfer distances and add an $L_1$ loss for the area of the light sources to encourage sharper lighting. Finally, for $\mathbf{Loss_{src}}$, we use $L_2$ loss for direction $\mathbf{d}$, $\log L_2$ loss for intensity $\mathbf{w}$ and bandwidth $\lambda$. More details are in the supplementary.  To train the shadow network, we use scale-invariant gradient loss proposed in \cite{niklaus20193d} and find that it leads to many fewer artifacts compared to a simple $L_2$ loss. We supervise indirect shading with $L_1$ loss and per-pixel lighting with rendering loss and $\log L_2$ loss similar to \cite{li2020inverse}.

\begin{figure}[t]
\centering
\includegraphics[width=\textwidth]{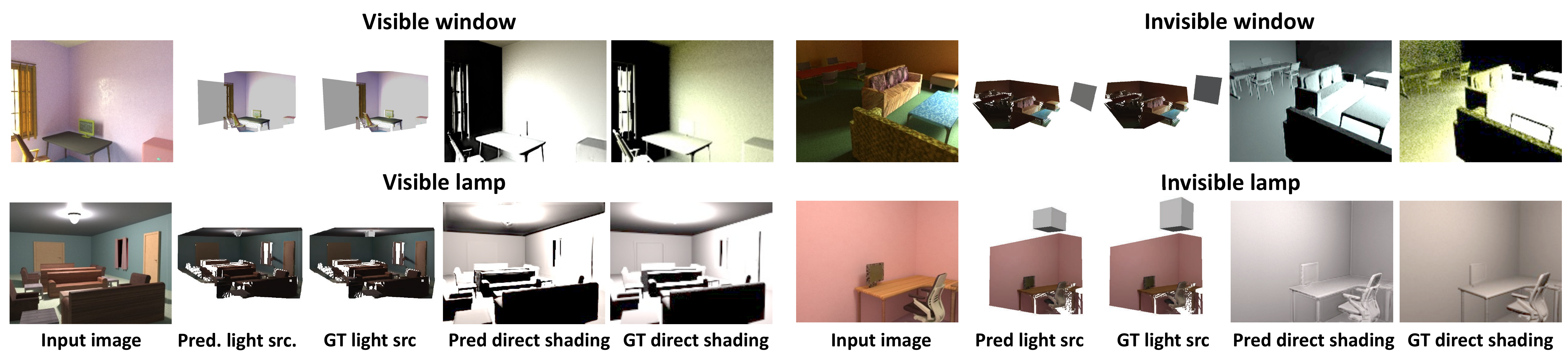}
\vspace{-0.8cm}
\caption{Light source prediction on the synthetic dataset for various types of light sources. We visualize the predicted and ground-truth light source geometry and their direct shading $\mathbf{E_j}$ without occlusion. Our method can recover both the geometry and radiance for the four types of light sources reasonably well, with $\mathbf{E_j}$ similar to the ground truths.}
\label{fig:synsrc}
\vspace{-0.3cm}
\end{figure}

\begin{figure*}[t]
\centering
\includegraphics[width=\textwidth]{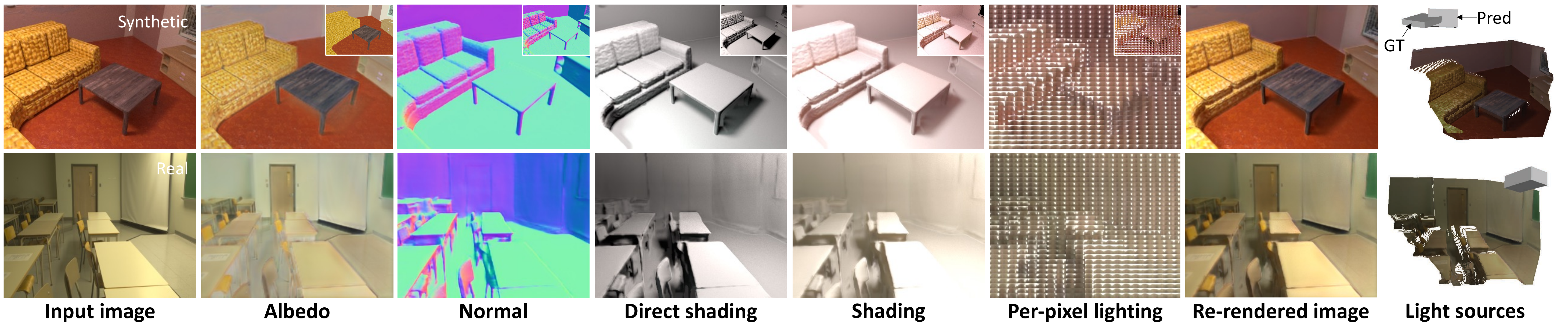}
\vspace{-0.8cm}
\caption{Our reflectance, geometry, lighting and rendering predictions on a synthetic and a real example. Ground truths for the synthetic example are shown in the insets. We observe that even for invisible light sources, our framework accurately reconstructs their geometry and intensities, which enables realistic rendering of the scene irradiance, shadows, interreflection and per-pixel lighting. }
\label{fig:prediction}
\vspace{-0.3cm}
\end{figure*}

\begin{figure}[t]
\centering
\begin{minipage}[c]{0.6\linewidth}
\includegraphics[width=\columnwidth]{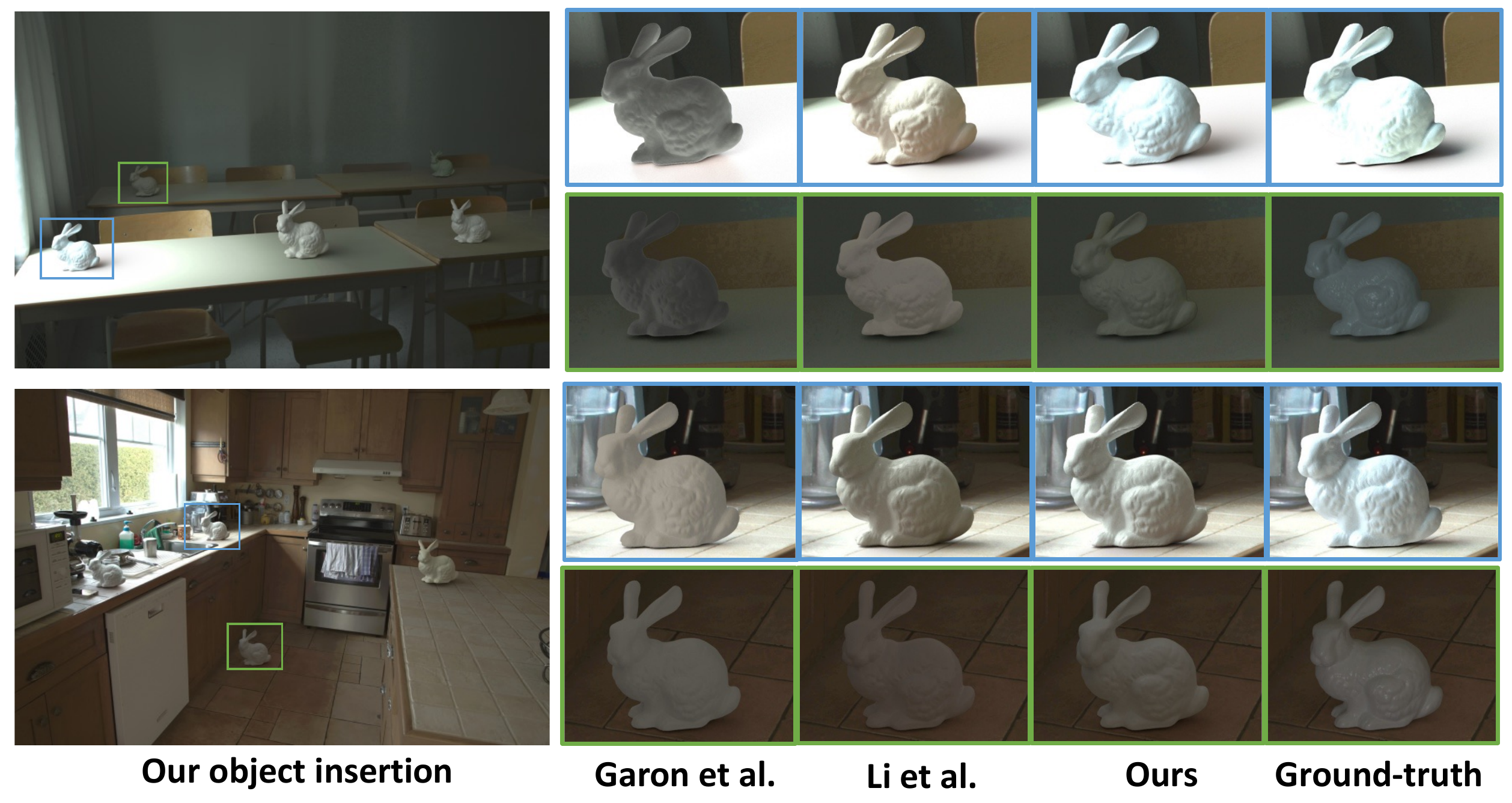}
\end{minipage}
\begin{minipage}[c]{0.38\linewidth}
\vspace{-0.4cm}
\caption{\mc{We achieve similar quality as prior state-of-the-art on Garon et al.~\cite{garon2019fast} dataset for object insertion on the surface. Our method accurately reconstructs the complex lighting from windows to render more realistic highlights and shadows. See Fig.~\ref{fig:garonPlus} for other editing tasks not possible for prior works.} }
\label{fig:garon}
\end{minipage}
\vspace{-0.3cm}
\end{figure}

\vspace{-0.3cm}
\paragraph{Training and inference} 
We use Adam \cite{adam} with learning rate $10^{-4}$ and $\beta$ (0.9, 0.999). We first train the $\mathbf{MNet}$, then we fix it and use its predictions as inputs to train the light source prediction networks  $\mathbf{InvLampNet}$,  $\mathbf{InvWinNet}$, $\mathbf{VisLampNet}$ and $\mathbf{VisWinNet}$ separately. We then train rendering modules independently by providing them with ground-truth $\mathbf{E_d}$ and $\mathbf{S}$ whenever they are required as inputs.  The typical inference time of our whole pipeline is less than 3s. More details are in the supplementary. 

\vspace{-0.3cm}
\paragraph{Refinement} While so far our framework can achieve high-quality light source prediction and indoor lighting editing in many cases, our differentiable neural rendering framework enables us to further refine the light source parameters by minimizing the rendering loss between the rendered image and the input image, leading to more robust and more realistic rendering. Fig. \ref{fig:optim} shows an example where we correct the intensity of an invisible lamp with our rendering loss-based refinement. Note that this is an extremely ill-posed problem.  A good initialization from our light source prediction networks is essential for the refinement to achieve good results. More discussions are in the supplementary. We only apply the refinement to real images shown in the paper, not to the synthetic images.

\begin{table}[t]
\begin{minipage}[c]{0.49\linewidth}
\centering
\begin{tabular}{|c|c|c|c|c|}
\hline 
& \multicolumn{2}{|c}{Geometry} & \multicolumn{2}{|c|}{Rendering} \\
\cline{2-5}
& \multicolumn{2}{|c|}{$\mathbf{Cham}(\mathbf{q_j}, \mathbf{\bar{q}_j})$} & \multicolumn{2}{c|}{$\mathbf{E_j}$} \\
\cline{2-5}
& Gt. & Pred. & Gt. & Pred. \\
\hline
Vis. lamp  & 0.279 & 1.15 & 0.317 & 0.557  \\
\hline
Vis. window & 0.415 & 1.14 & 0.849 & 0.952 \\
\hline
Inv. lamp & 0.712  & 0.988 &  0.289 & 0.357 \\
\hline 
Inv. window & 3.50 & 3.71 &  0.312 & 0.328  \\ 
\hline
\end{tabular}
\caption{Light source prediction on OpenRooms with ground truth and predicted depth. We report RMSE chamfer loss and $L_1$ error of direct shading w/o shadows $\mathbf{E_{j}}$. }
\label{tab:src}
\end{minipage}\hfill
\begin{minipage}[c]{0.50\linewidth}
\centering
\begin{tabular}{|c|c|c|c|c|c|}
\hline
 \multicolumn{2}{|c}{Direct shading} & \multicolumn{2}{|c}{Shading} & \multicolumn{2}{|c|}{Perpix. envmap}  \\
\multicolumn{2}{|c}{$\mathbf{E_d}$} & \multicolumn{2}{|c}{$\mathbf{E}$} & \multicolumn{2}{|c|}{$\mathbf{L}$} \\
\hline
Gt. & Pred. & Gt. & Pred. & Gt. & Pred. \\
\hline 
0.282 & 0.325 & 0.336 & 0.391 & 0.090 & 0.105  \\ 
\hline
\end{tabular}
\caption{Quantitative errors for our neural rendering framework on OpenRooms with ground-truth and predicted depth. We report $L_1$ loss for the sum of direct shading with shadows $\mathbf{E_d}$ and shading with global illumination $\mathbf{E}$. We report $\log L_2$ loss for per-pixel lighting $\mathbf{L}$. }
\label{tab:synrender}
\end{minipage}
\vspace{-0.3cm}
\end{table}

\begin{table}[t]
\centering
\begin{minipage}[c]{0.58\linewidth}
\small
\begin{tabular}{|c|c|c|c|}
\hline
Gardner et al. \cite{gardner2017indoor} & Garon et al. \cite{garon2019fast} & Li et al. \cite{li2020openrooms}  \\
\hline 
 72.4\% & 69.2\% & 52.0\% \\
\hline
\end{tabular}
\end{minipage}
\begin{minipage}[c]{0.40\linewidth}
\caption{User study on Garon et al. dataset. We require 200 users to compare our results with prior results and report the percentage of users who believes ours are better.}
\label{tab:garonQuan}
\end{minipage}
\vspace{-0.4cm}
\end{table}
\vspace{-0.2cm}
\section{Experiments}
\label{sec:experiment}
\vspace{-0.2cm}

\begin{figure*}[t]
\centering
\includegraphics[width=\textwidth]{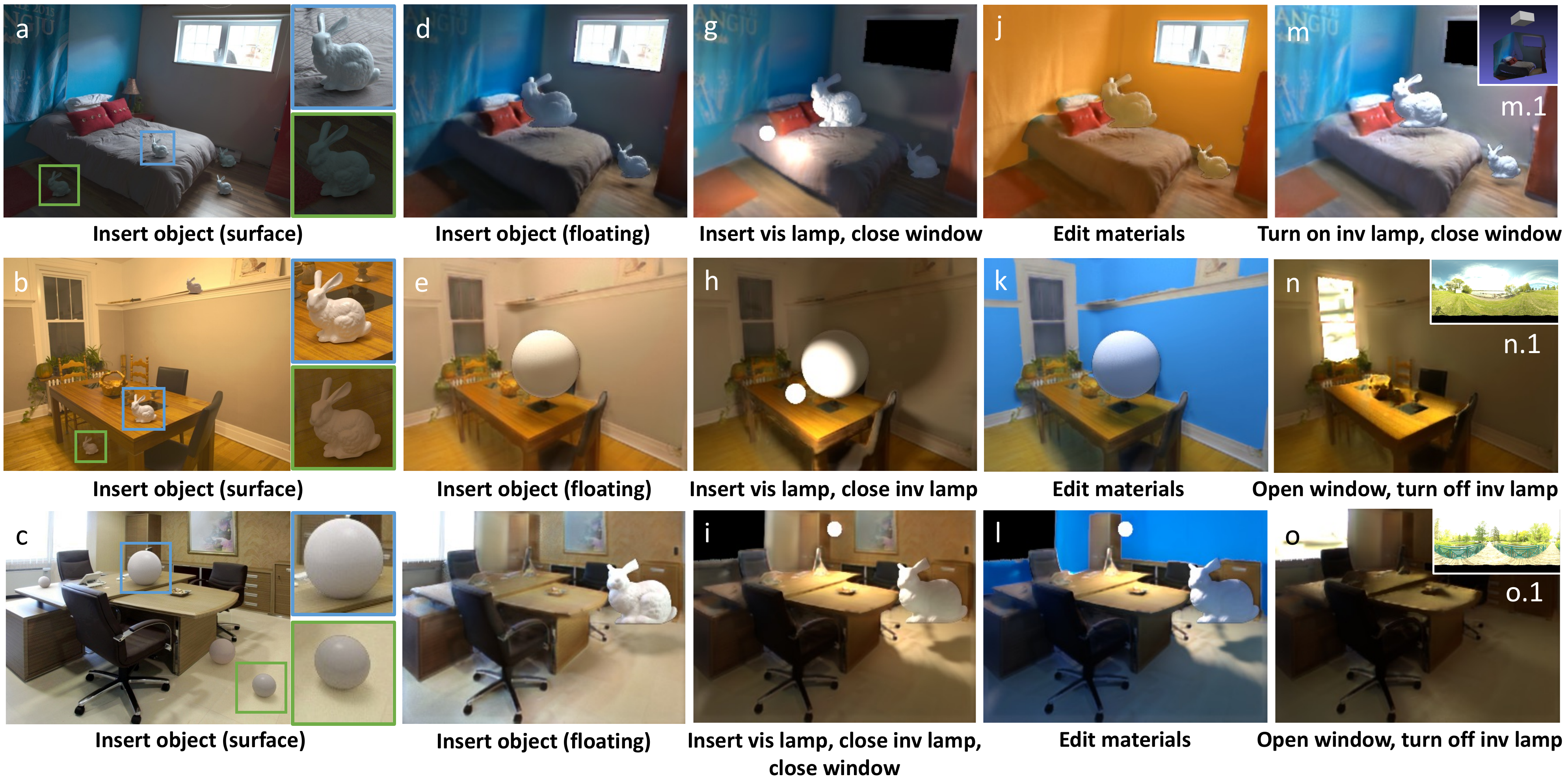}
\vspace{-0.8cm}
\caption{Various editing applications demonstrated on 3 real examples. In addition to high-quality object insertion results (a, b and c), our light source representations combined with our neural rendering framework allows us to edit geometry, material and lighting of indoor scenes with non-local effects being effectively modeled. This includes distant shadows projected to the bed, table and floor (d, e, f and i) or to the entire room when the object blocks the light source (g and h), changing color of walls that causes non-local color bleeding (j, k and l) and adding virtual light sources into the scene (g, h, i, l, m, n, o), including turning on a lamp or opening a virtual window. }
\label{fig:garonPlus}
\vspace{-0.3cm}
\end{figure*}

\begin{figure}[t]
\centering
\begin{minipage}[c]{0.6\linewidth}
\includegraphics[width=\columnwidth]{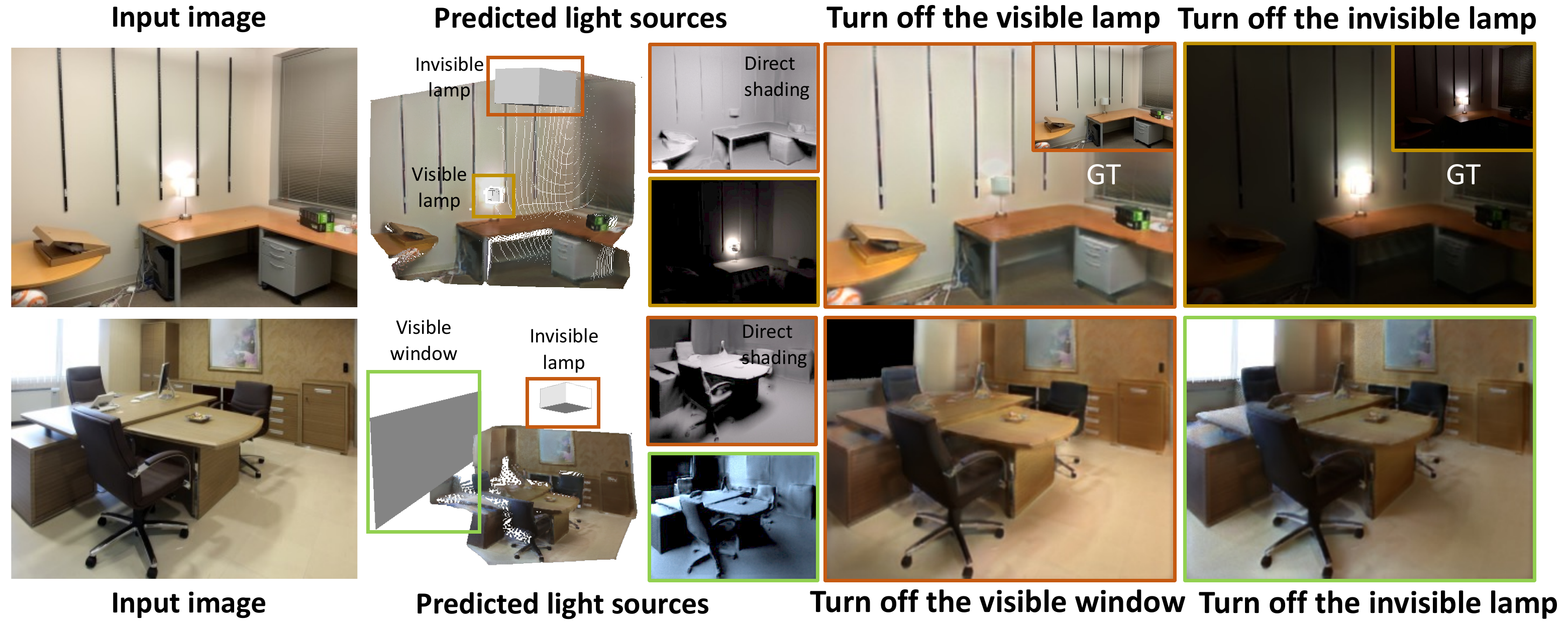}
\end{minipage}\hfill
\begin{minipage}[c]{0.38\linewidth}
\caption{Our accurate reconstruction of visible/invisible light sources allows separating their contributions and turn them on and off. Our results closely match the ground-truth insets.}
\label{fig:onOff}
\end{minipage}
\vspace{-0.5cm}
\end{figure}

\zl{We will present light source estimation and neural rendering results on real and synthetic data, as well as various scene editing applications, especially light editing, on real data.} For synthetic data, \zlnew{we test both ground-truth depths and predicted depths from DPT \cite{ranftl2021vision} w/o fine-tuning and use ground-truth light segmentation masks. } Synthetic qualitative results in the main paper are with ground-truth depth. We include qualitative comparisons with predicted depth in the supplementary. For real data, we generate all depth predictions using DPT \cite{ranftl2021vision} and manually label the segmentation masks. While not being our focus, \zl{we also evaluate a Mask RCNN \cite{he2017mask} for light source detection in the supplementary}. 

\vspace{-0.3cm}
\paragraph{Light source predictions and neural rendering} \zl{Fig.~\ref{fig:synsrc} shows qualitative results of our light source predictions on the synthetic dataset. For each of the four types of light sources, we pick up one from the testing set to visualize their predicted and ground-truth geometry and their rendered direct shading $\mathbf{E_j}$ without occlusion. We observe that our method can recover both the geometry and radiance for all 4 types of light sources reasonably well, which enables us to render their direct shading quite close to the ground-truths. The main errors in our rendered direct shading are caused by global shifts of colors and intensities, while the locations of highlights are usually correct. This is reasonable given the scale ambiguity between materials and lighting. Tab.~\ref{tab:src} summarizes the quantitative errors. We see that the errors for windows are larger than those of lamps, since the outdoor lighting coming through windows is much more complicated compared to area lighting. We also observe that the direct shading errors for invisible light sources are lower. This is because their overall contributions are usually lower since many of them are far away from the camera location. } We also report quantitative numbers using predicted depths from DPT without fine-tuning. We observe that our method achieves comparable rendering errors even with predicted depth. We include qualitative comparisons in the supplementary.

Fig. \ref{fig:prediction} shows a complete set of our neural rendering results on a synthetic and a real example. Quantitative results for neural rendering are summarized in Table \ref{tab:synrender}. For the synthetic example, our shadow prediction network combined with Monte-Carlo ray tracing allows for rendering distant shadows from a single depth map without boundary artifacts. Our indirect shading prediction network models non-local interreflections from only partial observations of geometry and materials. All the predictions combined together lead to accurate reconstruction of shading and per-pixel lighting. For the real example, even though we do not have ground-truths, we observe that the light source position, the highlight in the direct shading and shadows are all visually consistent. The re-rendered image with appearance closely matching the input image further demonstrates that our framework can generalize well to real examples. 

\vspace{-0.3cm}
\paragraph{Comparisons with prior works} \zl{We reiterate that our method enables applications (e.g. light source editing) that are not possible with any prior work. While this makes direct comparisons challenging, we compare on a subset of tasks like object insertion that previous work can support.} We use Garon et al. dataset for comparison, which is a widely-used, real dataset for spatially-varying lighting evaluation. Even though we are solving a harder problem, both qualitative and quantitative results in Fig.~\ref{fig:garon} and Tab.~\ref{tab:garonQuan} show that our method achieves performance comparable to the prior state-of-the-arts which only handle local editing of the scene. Our per-pixel lighting prediction can be used to render specular objects realistically, with highlights, shadows and spatial consistency being correctly modeled as shown in Fig.~\ref{fig:garon} and \ref{fig:garonPlus}. Specifically, our 3 SG sunlight representation and MIS based rendering layer allow us to better handle high-frequency, complex sunlight coming from the window, leading to rendering results closer to the ground-truths, as presented in Fig.~\ref{fig:garon}. 

\vspace{-0.3cm}
\paragraph{Novel scene editing applications} In addition to high-quality object insertion (a, b, c) with local highlights and shadows, the true advantage of our framework is its ability to handle non-local effects in scene editing application, which is only made possible by our accurate reconstruction of indoor light sources and high-quality neural rendering that models multiple complex light transport effects, such as hard/soft shadows, interreflection and directional lighting. These non-local effects include distant shadows and highlights, which is shown in (d, e, f) of Fig.~\ref{fig:garonPlus} where the inserted virtual objects block the sunlight coming from the visible window or the light from the invisible lamp, projecting shadows to the bed, table and floor respectively. This is further demonstrated in (g, h, i), where the inserted virtual lamp causes highlights on the surface of nearby geometry and causes shadows that cover the whole wall behind the inserted virtual bunny and sphere. Moreover, our framework also allows non-local interreflection to be accurately modified. As shown in (j, k, l), as we change the color of walls to orange and blue, our indirect shading network paints the inserted white objects with correct color bleeding. In (m, n, o), we demonstrate our framework's ability to turn on an invisible lamp or open a virtual window. Note that in n, o, we use the 3 SG approximation of the environment map shown in n.1 and o.1 respectively. Our representation combined with our neural renderer can render realistic sunlight. 

Our accurate reconstruction of indoor light sources further allows us to separate their contributions. As shown in both Fig.~\ref{fig:teaser} and \ref{fig:onOff}, our framework allows us to turn off visible and invisible lamps or windows in the scene, with changed appearance similar to the ground-truth insets\footnote{The second example is from the internet so we do not have its ground truth.}.

\vspace{-0.2cm}
\section{Conclusions}
\label{sec:conclusions}
\vspace{-0.2cm}

We presented a method that enables full indoor scene relighting and other editing operations from a single LDR image with its predicted depth and light source segmentation mask. A key innovation in our solution is our lighting representation; we estimate multiple global 3D parametric lights (lamps and windows), both visible and invisible. A second important component is our hybrid renderer, capable of producing high-quality images from our scene representation using a combination of Monte Carlo and neural techniques. We show that this careful combination of an editable lighting representation and neural rendering can handle challenging scene editing applications including object insertion, material editing, light source insertion and editing, with realistic global effects.

\vspace{-0.3cm}
\paragraph{Acknowledgements} We thank NSF CAREER 1751365, 2110409, 1703957 and CHASE-CI, ONR N000142012529, N000141912293, a Google Award, generous gifts from Adobe, Ron L. Graham Chair and UCSD Center for Visual Computing and Qualcomm Innovation Fellowship

\appendix

\section{Overview of the Supplementary Material}
The supplementary material is organized as follows:
\begin{tight_itemize}
\item A \href{https://drive.google.com/file/d/1IuMuJ4QyVGIWNhN_HhwOhyJuQ7Ud26b5/view?usp=sharing}{video} to demonstrate the consistency of our scene editing results (Sec. \ref{sec:sup_video} )
\item Network and training details (Sec.~\ref{sec:sup_network})
\item More qualitative and quantitative light source prediction and neural rendering results on our synthetic dataset (Sec.~\ref{sec:sup_exp})
\item Light editing results with predicted masks (Sec.~\ref{sec:sup_lightEdit}). 
\item Limitations and future works (Sec.~\ref{sec:sup_limitation}). 
\end{tight_itemize}

\section{Video}
\label{sec:sup_video}

We include a \href{https://drive.google.com/file/d/1IuMuJ4QyVGIWNhN_HhwOhyJuQ7Ud26b5/view?usp=sharing}{video} covering various scene editing applications. In the \href{https://drive.google.com/file/d/1IuMuJ4QyVGIWNhN_HhwOhyJuQ7Ud26b5/view?usp=sharing}{video}, we move virtual light sources and objects to change highlights and shadows, and gradually modify the wall color to edit global illumination. Even though we do not explicitly add any smoothness constraint, our framework manages to achieve consistent scene editing results. This is probably because our framework explicitly follows the physics of the image formation process, which provides a natural regularization for our rendering results to be consistent.  
\section{Network Architecture and Training Details}
\label{sec:sup_network}

We train our network on the OpenRooms Dataset \cite{li2020openrooms}, which is the only dataset that provides all the necessary ground truths for training our light source estimation and neural rendering frameworks,  including depth $\mathbf{D}$, SVBRDF (diffuse albedo $\mathbf{A}$, normal $\mathbf{N}$ and roughness $\mathbf{R}$), direct shading $\mathbf{E}_{\mathcal{L}, \mathcal{W}}$ and shadow $\mathbf{S}_{\mathcal{L},\mathcal{W}}$ for each individual light source, shading with indirect illumination $\mathbf{E}$ and per-pixel lighting $\mathbf{L}$. The dataset contains 1287 indoor scenes, with 118,233 images in total. We utilize 108,159 images rendered from 1178 scenes for training and the rest for testing. All the synthetic results are generated from the testing set. The comprehensive supervision provided by the OpenRooms dataset \cite{li2020openrooms} allows us to train each module of our framework separately, which greatly simplifies the training process. The number of iterations for training each network and batch sizes are summarized in Tab.~\ref{tab:sup_train}.

For all network figures in the supplementary, $CX_1$-$KX_2$-$SX_3$-$PX_4$-$GX_5$ represents a convolutional layer with $X_1$ channels, kernel size $X_2$, stride size $X_3$, padding size $X_4$, followed by a group normalization layer with $X_5$ channels per group. $Up$ represents a bilinear interpolation layer that doubles the resolution of the input feature map.

\subsection{Material Prediction} 
\begin{figure}[t]
\centering
\begin{minipage}[c]{0.6\textwidth}
\includegraphics[width=\columnwidth]{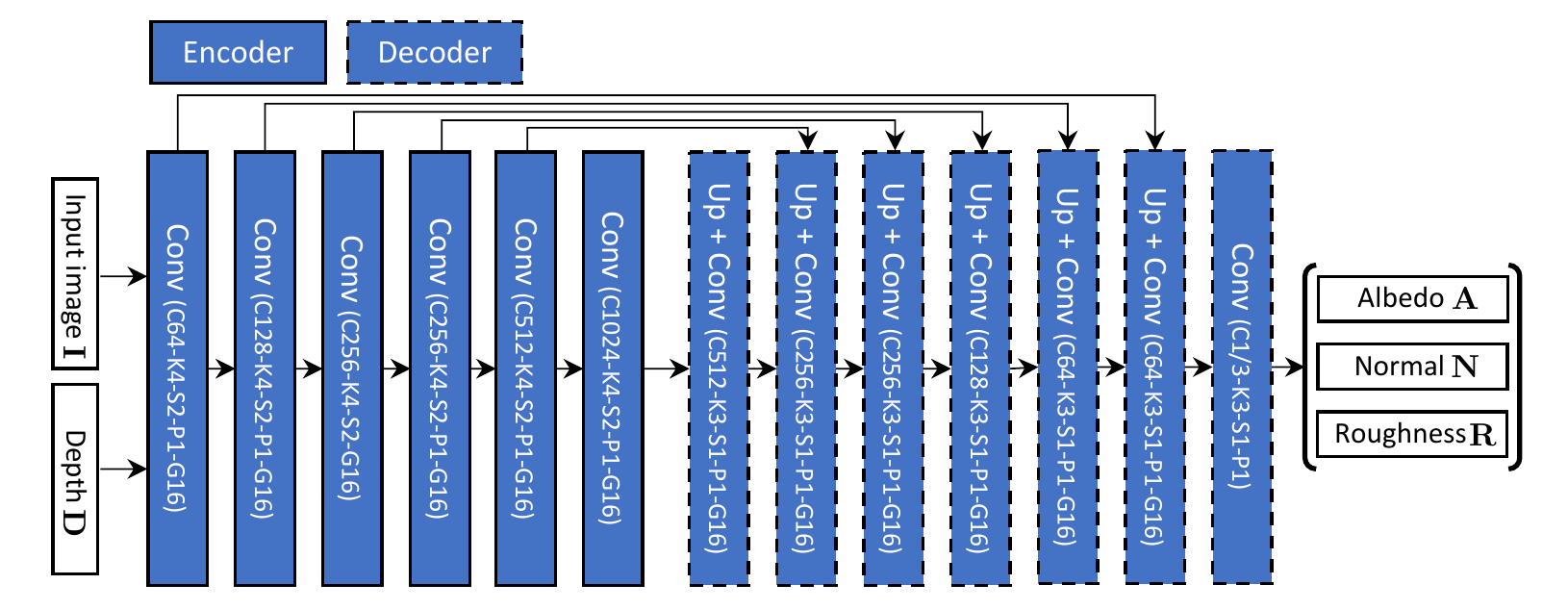}
\end{minipage}\hfill
\begin{minipage}[c]{0.38\textwidth}
\caption{Network architecture of $\mathbf{MNet}$ for material parameter prediction.}
\label{fig:sup_mnet}
\end{minipage}
\vspace{-0.2cm}
\end{figure}

We first train $\mathbf{MNet}$ for SVBRDF prediction. The network architecture is shown in Fig.~\ref{fig:sup_mnet}. The input to the network is a $240\times 320$ LDR image $\mathbf{I}$ and the depth map $\mathbf{D}$, while the outputs are diffuse albedo $\mathbf{A}$, normal $\mathbf{N}$ and roughness $\mathbf{R}$. Similar to prior work \cite{li2020inverse}, we use three decoders but one shared encoder to predict three BRDF parameters because these parameters are correlated. We add skip-links to help reconstruct details. The loss function is the sum of $L_2$ loss on the three BRDF parameters. 
\begin{equation}
\small
\sum_{\mathbf{X\in\{A, N, R\}} } (\mathbf{X - \bar{X}})^2.
\end{equation}
Note that we normalize the ground-truth $\mathbf{D}$ so that its mean is equal to 3 before we send it to every network. The reason is that there is a scale ambiguity for single image depth prediction using DPT network \cite{ranftl2021vision}. We find that this is important for the networks to generalize well to real images.  

\begin{figure}[t]
\centering
\begin{minipage}[c]{0.6\textwidth}
\includegraphics[width=\columnwidth]{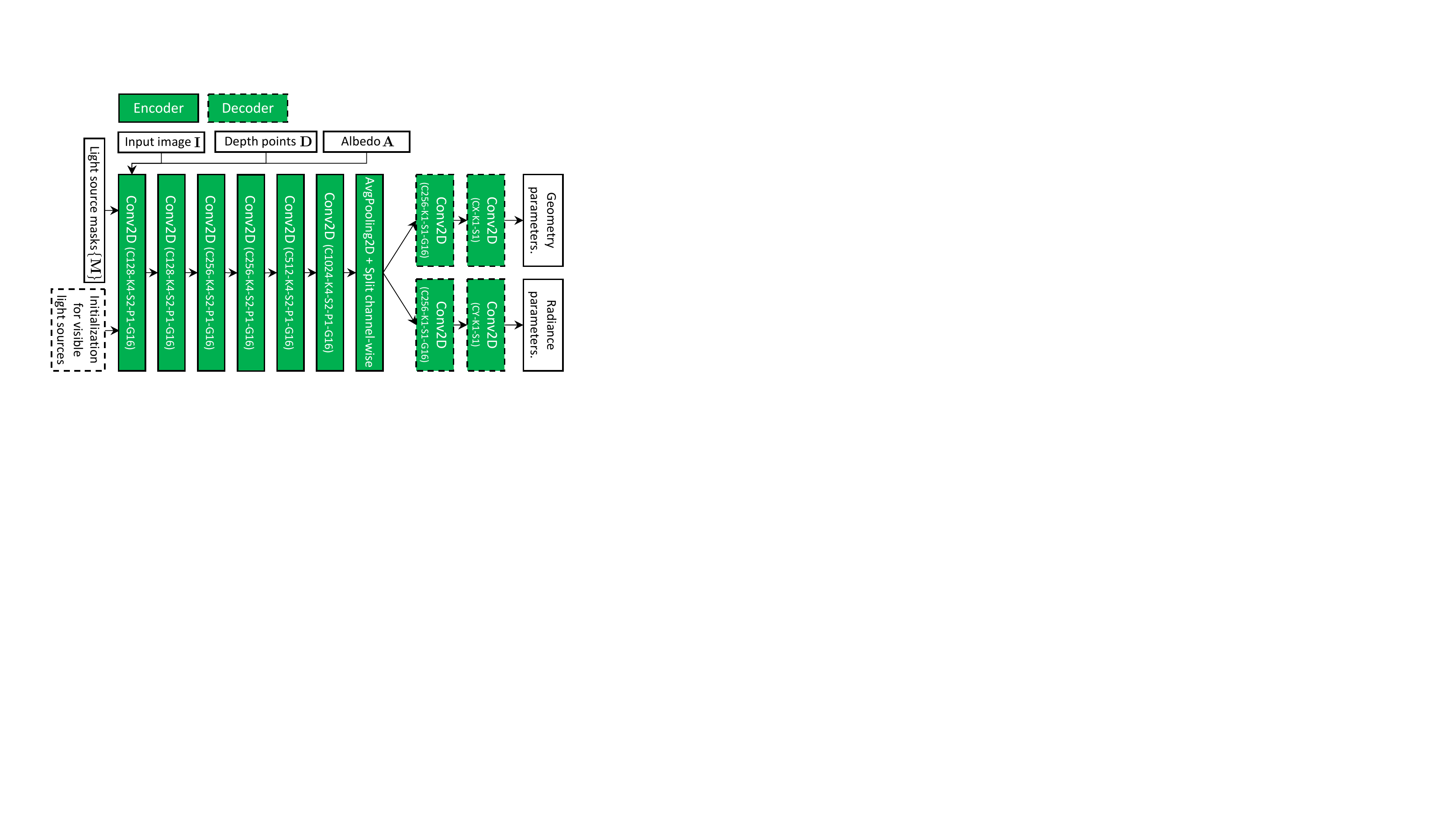}
\end{minipage}\hfill
\begin{minipage}[c]{0.38\textwidth}
\caption{Network architecture for light source prediction. All four light source prediction networks share a similar architecture. Networks for visible light source prediction have initialization of light source geometry as inputs. The numbers of output channels differ according to the light source type. }
\label{fig:sup_lightSrcNet}
\end{minipage}
\vspace{-0.2cm}
\end{figure}

\begin{figure}[!!t]
  \footnotesize
  \centering
    \begin{minipage}[t]{0.3\linewidth}
      \raisebox{-3.0cm}
               {
                \includegraphics[width=\linewidth]{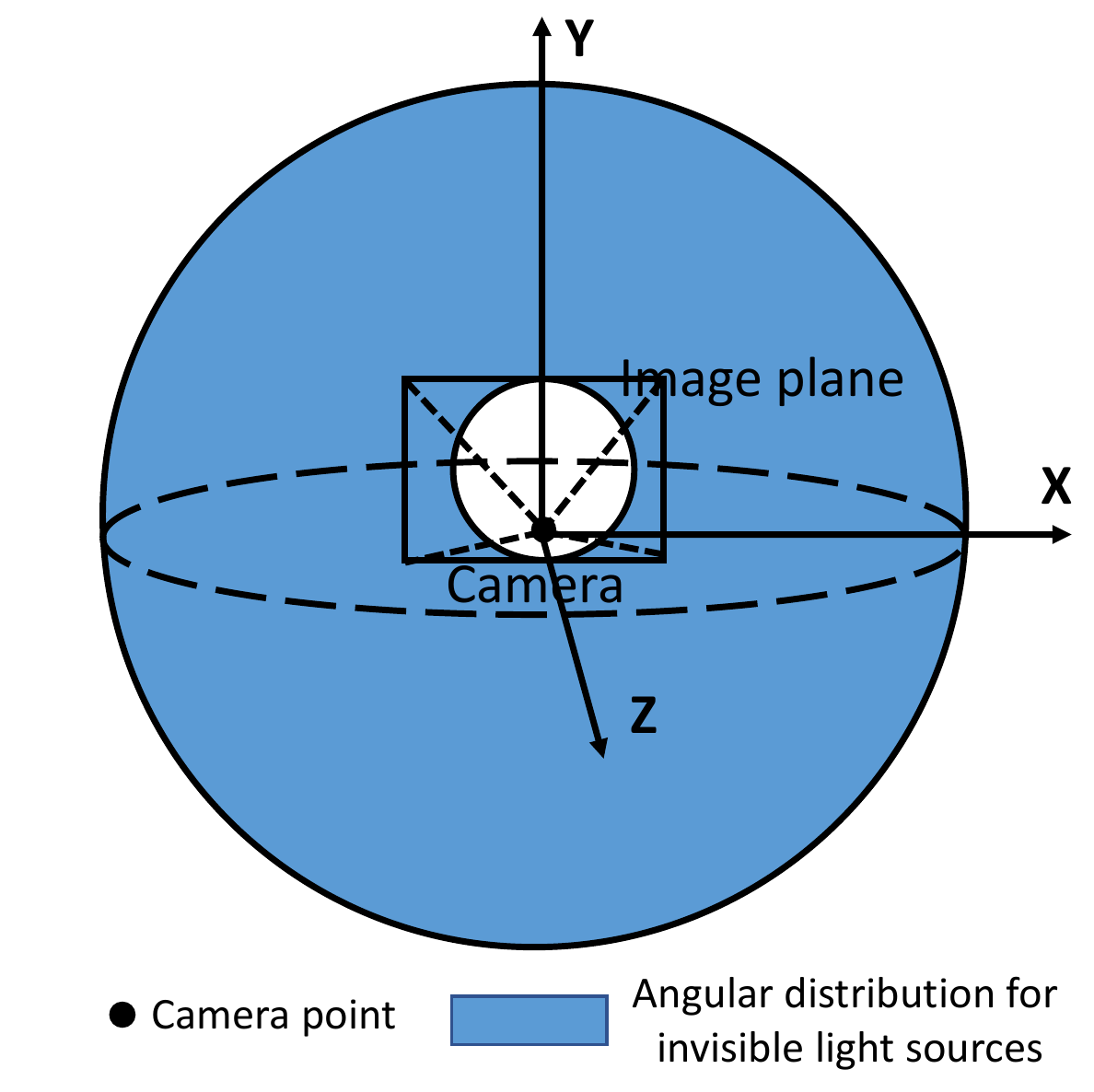}    
               }
    \end{minipage}\hfill
    \begin{minipage}[t]{0.65\linewidth}
      \caption{
        Visualization of parameterization for the centers of invisible light sources. Our parameterization encourages centers of invisible light sources to be outside the camera frustum. 
        \label{fig:sup_invPara}
      }
    \end{minipage}
\vspace{-0.2cm}
\end{figure}

\subsection{Light Source Prediction} 
Once we finish training the $\mathbf{MNet}$, we use its predictions $\mathbf{A}$, $\mathbf{N}$ and $\mathbf{R}$ as inputs, combined with image $\mathbf{I}$, depth map $\mathbf{D}$ and light source segmentation mask $\mathbf{M_j}$ to train our light source prediction networks. We have four types of light source prediction networks for four types of light sources, \{visible/invisible\} and \{lamp/window\}, which share similar architectures but differ in parameterization and inputs. As shown in Fig.~\ref{fig:sup_lightSrcNet}, the encoder architecture consists of six 2D convolutional layers with stride 2. We also tried to use pre-trained ResNet-18 \cite{he2016deep} but the results were worse.

\vspace{-0.4cm}
\paragraph{Invisible light sources prediction}
The inputs to invisible window and invisible lamp prediction networks are exactly the same, which include the LDR image ($\mathbf{I}$), depth ($\mathbf{D}$), albedo ($\mathbf{A}$) and the sum of light source masks $\mathbf{M} = \sum_{\mathbf{j}}\mathbf{M_j}$. We first project 1-channel depth map $\mathbf{D}$ into 3-channel point cloud before we concatenate it with other inputs. The radiance of a lamp is controlled by intensity $\mathbf{w}$ only. We use $\tan$ function to project the initial network output $\mathbf{\tilde{w}}$ in range $[0, 1]$ to high dynamic range 
\begin{equation}
\mathbf{w} = \tan(\frac{\pi}{2}\mathbf{\tilde{w}} ). 
\end{equation}
The radiance of a window is controlled by 3 SGs, which correspond to sun, sky and ground of outdoor illumination. We encourage the SG corresponding to sun to represent high-frequency directional lighting. Therefore, we introduce two more parameters $\lambda^{\max}$ and $\lambda^{\min}$ for each SG to constrain their bandwidth parameters. Similarly, let $\tilde{\mathbf{w}}_{\mathbf{k}}$, $\mathbf{\tilde{d}_{k}}$ and $\tilde{\lambda}_\mathbf{k}$ be the initial predictions, where $\mathbf{\tilde{d}}_k$ is in the range of [-1, 1], $\tilde{\mathbf{w}}_{\mathbf{k}}$ and $\tilde{\lambda}_k$  are in the range of $[0, 1]$, we have 
\begin{eqnarray}
\mathbf{w_k} &=& \tan(\frac{\pi}{2}\mathbf{\tilde{w}_k} ) \\
\mathbf{d_k} &=& \mathtt{normalize}(\mathbf{\tilde{d}_k} ) \\
\lambda_\mathbf{k} &=& \tan(\frac{\pi}{2} (\tilde{\lambda}_{\mathbf{k}}(\lambda^{\max}_{\mathbf{k}} - \lambda^{\min}_{\mathbf{k}}) + \lambda^{\min}_{\mathbf{k}} ) )
\end{eqnarray}
The value of $\lambda^{\max}$ and $\lambda^{\min}$ parameters for sun, sky and ground are summarized in Tab.~\ref{tab:sup_lambda}. 

\begin{table}[t]
\centering
\begin{tabular}{|c|c|c|c|}
\hline
& sun & sky & ground  \\
\hline
($\lambda^{\min}$, $\lambda^{\max}$) & (0.9, 1-$10^{-6}$)  & (0, 1-$10^{-4}$)  & (0, 1-$10^{-4}$) \\
\hline
\end{tabular}
\caption{The value of $\lambda^{\min}$ and $\lambda^{\max}$ for SGs correspond to sun, sky and ground respectively. $\lambda^{\max}$ is set slightly less than $1$.}
\vspace{-0.6cm}
\label{tab:sup_lambda}
\end{table}

We represent the geometry of the invisible window and invisible lamp using a plane $\{\mathbf{c, x, y}\}$ and a 3D box $\{\mathbf{c, x, y, z}\}$ respectively, where $\mathbf{c}$ is the center and $\{\mathbf{x, y, z}\}$ are the three axes. To recover axes for lamps, we predict Euler angles $\alpha$, $\beta$ and $\gamma$  as well as the axis length $l_\mathbf{x}$, $l_\mathbf{y}$ and $l_\mathbf{z}$, which combined together can be used to compute $\mathbf{x,y,z}$. To recover axes for windows,  we first predict  initial axes $\tilde{\mathbf{y}}$ and $\mathbf{z}$ that are perpendicular to each other, as well as axis lengths $l_{\mathbf{x}}$ and $l_{\mathbf{y}}$. Let $\mathbf{u}=[0, 1, 0]$ be the up vector. The final axis predictions are computed as 
\begin{eqnarray}
\mathbf{y} &=& \mathtt{normalize}(\tilde{\mathbf{y}} + \mathbf{u}) l_{\mathbf{y}} \\
\mathbf{x} &=& \mathtt{normalize}(\mathtt{cross}(\mathbf{z, y}) ) l_{\mathbf{x}}
\end{eqnarray}
The intuition of introducing $\mathbf{u}$ is that for most windows, their $\mathbf{y}$ is close to up vector $\mathbf{u}$. Therefore, we predict the difference between $\mathbf{y}$ and $\mathbf{u}$ to make the training easier. 

To predict centers of invisible light sources, we notice that if we directly output center $\mathbf{c}$, in some cases, the invisible light sources will be located inside the camera frustum or even across scene geometry, causing artifacts in the rendering. Thus, we change the parameterization to push their centers outside the camera frustum, as visualized in Fig.~\ref{fig:sup_invPara}. We decompose the center $\mathbf{c}$ into direction $\mathbf{d_c}$ and length $l_{\mathbf{c}}$. Let $f$ be the field of view for the short axis of the image plane, and $\mathbf{x_{c}}$, $\mathbf{y_c}$ and $\mathbf{z_{c}}$ be the camera coordinate system as shown in Fig.~\ref{fig:sup_invPara}. We first predict $\theta_{\mathbf{c}} \in [0, \pi - f]$, $\phi_{\mathbf{c}} \in [-\pi, \pi]$ and $l_{\mathbf{c}}\in[0, \infty]$. Then, we compute center $\mathbf{c}$:
\begin{eqnarray}
\mathbf{d_{c}} &=& \mathbf{x_c}\sin\theta_{\mathbf{c}}\cos\phi_{\mathbf{c}} + \mathbf{y_{c}}\sin \theta_{\mathbf{c}}\cos\phi_{\mathbf{c}} \\
&& + ~~\mathbf{z_c}\cos\theta_\mathbf{c} \\
\mathbf{c} &=& \mathbf{d_c} l_{\mathbf{c}}
\end{eqnarray}

\vspace{-0.6cm}
\paragraph{Visible window prediction} Both geometry and radiance representations for visible windows are the same as those of invisible windows. However, for visible windows, instead of directly predicting their location from a single image, we first compute their initial light source center $\mathbf{c^{init}}$ based on their segmentation masks $\mathbf{M}_{\mathcal{W}}$ and depth $\mathbf{D}$, send the initial results to the network and then predict the difference between the initial estimation and the ground-truths. More specifically, we define $[\mathbf{X}; \mathbf{Y}; -\mathbf{1}]$ to be a 3-channel image representing pixel locations on the image plane. We introduce function $\mathbf{Edge}(\mathbf{M}, n)$ to compute the edge pixels of mask $\mathbf{M}$, where 
\begin{equation}
\small
\mathbf{Edge}(\mathbf{M}, n) = \mathtt{dilation}(\mathbf{M}, n) - \mathbf{M}.
\end{equation}
Formally, the initial center $\mathbf{c^{init}}$ is defined as
\begin{eqnarray*}
\small
\mathbf{c^{init}} &=& \mathtt{mean}([\mathbf{X}; \mathbf{Y}; -\mathbf{1}] \mathbf{M}_{\mathcal{W}}, \mathtt{axis}=3) \\
\small
&& \mathtt{mean}(\mathbf{DEdge}(\mathbf{M}_{\mathcal{W}}, 7 ) )
\end{eqnarray*}
$\mathbf{c^{init}}$ is then sent to the network by concatenating with other inputs as an extra 3-channel image. Let $\tilde{\mathbf{c}}$ be the output from the network. The final prediction $\mathbf{c}$ is defined as
\begin{equation}
\small
\mathbf{c} = \mathbf{c^{init}} + \mathbf{\tilde{c}}
\end{equation}

\vspace{-0.4cm}
\paragraph{Visible lamp prediction} The radiance of a visible lamp is simply represented by intensity $\mathbf{w}$. As shown in Fig.~\ref{fig:lampRep} in the main paper, the geometry of a visible lamp is represented by reflecting the visible surface with respect to the center to compute the invisible and boundary area. Similar to the visible window case, we first compute the initial center as
\begin{eqnarray*}
\small
\mathbf{c^{init}} &=& \mathtt{mean}([\mathbf{X}; \mathbf{Y}; -\mathbf{1}] \mathbf{M}_{\mathcal{W}}, \mathtt{axis}=3) \\
&& \mathtt{mean}(\mathbf{D}\mathbf{M}_{\mathcal{W}} )
\end{eqnarray*}
Since lamps are usually small and errors in their geometry prediction can cause highlight artifacts, we add a stronger regularization by requiring that the final predicted center $\mathbf{c}$ must be in the same camera ray of the initial center $\mathbf{c^{init}}$. We decompose $\mathbf{c^{init}}$ into direction $\mathbf{d_{c}^{init}}$ and length $l_{\mathbf{c}}^{\mathbf{init}}$ and predict $\tilde{l}_{\mathbf{c}}$ so that we have 
\begin{equation}
\small
\mathbf{c} = \mathbf{d_{c}^{init}}(l_{\mathbf{c}}^{\mathbf{init}}+ \tilde{l}_{\mathbf{c}})
\end{equation}
Once we get the center $\mathbf{c}$, we can compute invisible area and boundary area based on that. Let $\mathbf{q}$ be one point on the visible area of a lamp, and $\mathbf{N(q)}$ be its normal. Let $H$ be the height of the image. Recall that $f$ is the field of view for the $\mathbf{y}$ axis. The area of $\mathbf{p}$ can be computed as
\begin{equation}
\small
\mathtt{area}(\mathbf{q}) = \left(\frac{1}{H}\tan(\frac{f}{2})\right)^{2} \frac{1}{\max(\mathbf{N(q)}^{T}\cdot[0;0;1], 0) } 
\label{eq:sup_areaVis}
\end{equation}
The corresponding invisible area $\mathbf{\hat{q}}$ can be computed as 
\begin{eqnarray}
\small
\mathbf{\hat{q}} &=& 2 (\mathbf{c - (q\cdot d_c)d_c }) + \mathbf{q} \label{eq:sup_pInv}\\
\mathtt{area}(\mathbf{\hat{q}}) &=&  \mathtt{area}(\mathbf{q}) \label{eq:sup_areaInv}\\
\mathbf{N(\hat{q})} &=& \mathbf{N(q)} - 2(\mathbf{N(q)}\cdot \mathbf{d_c}) \mathbf{d_c} \label{eq:sup_nInv}
\end{eqnarray}
As for edge pixels, they can be computed as
\begin{equation}
\small
\mathbf{Edge}(\mathbf{M}_{\mathcal{L}}, -1) = \mathbf{M}_{\mathcal{L}} - \mathtt{erosion}(\mathbf{M}_{\mathcal{L}}, 1)
\end{equation}
We create edge surface with center $\mathbf{q_e}$ so that
\begin{eqnarray}
\small
\mathbf{q_e} &=& \frac{\mathbf{q} + \mathbf{\hat{q}}}{2} \label{eq:sup_pEdge}\\
\mathtt{area}(\mathbf{q_e}) &=& ||\mathbf{q - \hat{q}}|| \left(\frac{1}{H}\tan(\frac{f}{2})\right) \label{eq:sup_areaEdge} \\
\mathbf{N(q_e)} &=& \mathtt{normalize}(\mathbf{q_e} - \mathbf{c}) \label{eq:sup_nEdge}
\end{eqnarray}
Note that both $\mathbf{\hat{q}}$ and $\mathbf{q_e}$ are differentiable with respect to predicted center $\mathbf{c}$, which allows us to supervise the center prediction using the chamfer distance loss, which will be discussed below. 

\vspace{-0.4cm}
\paragraph{Loss functions} We train the networks for visible lamps prediction using two loss functions $\mathbf{Loss_{ren}^{j}}$ and $\mathbf{Loss_{geo}^{j}}$. The $\mathbf{Loss_{ren}^{j}}$ is the $L_{1}$ loss between the rendered direct shading and the ground-truth of light source $\mathbf{j}$
\begin{equation*}
\small
\mathbf{Loss_{ren}^{j}} = |\mathbf{E_j} - \mathbf{\bar{E}_j} |.
\end{equation*}
We also considered $\log L_2$ loss but find that it focuses too much on low intensity regions and can cause wrong highlights. The geometry loss has two parts, a RMSE Chamfer distance loss and $L_1$ area loss. To compute the RMSE Chamfer distance loss, we randomly sample points on the surface of lamps and compute their RMSE Chamfer distance between the points sampled from ground-truth geometry. We find that compared to standard $L_2$ Chamfer distance, RMSE Chamfer distance can make the training more stable, especially for invisible light sources. To compute the $L_1$ area loss, we compute the sum of surface area of the predicted lamp and the ground-truth lamp. We observe that the area loss is important in preventing the network from predicting too large light sources, which may cause shadows to be blurry. 
\begin{equation*}
\small
\mathbf{Loss_{geo}^{j}} \!\!\! = \mathbf{Cham}(\{\mathbf{q_j}\}, \{\mathbf{\bar{q}_j}\}) + \omega_{a}|\mathtt{area}(\mathbf{j}) - \mathtt{area}(\bar{\mathbf{j}})|,
\end{equation*}
where $\omega_a$ is equal to 0.8. 

For visible windows, we use the same $\mathbf{Loss^{j}_{ren}}$ and $\mathbf{Loss^{j}_{geo}}$ but also add direct supervision on the light source spherical Gaussian parameters. The way we compute the ground-truth window radiance parameters is introduced in Sec.~\ref{sec:sup_gtLightSrc}. In summary, we use $L_2$ loss to supervise direction $\mathbf{d}$ and $\log L_2$ loss to supervise intensity $\mathbf{w}$ and bandwidth $\lambda$.
\begin{eqnarray}
\small
\!\!\!\!\!\!\!\!\!\!&&\!\!\!\!\!\!\!\!\!\mathbf{Loss_{src}^{\mathcal{W}}} =\!\!\!\!\!\!\!\!\! \sum_{k}^{\text{sun, sky, grd} } \!\!\!\!\!\!\!\! \omega_{\mathbf{k}} \Big(\omega_{\mathbf{w}}||\log(\mathbf{w_k}\!+\!1)\! -\!  \log(\mathbf{\bar{w}_k}\!+\!1)||^2  \label{eq:sup_srcLoss}\\
\!\!\!\!\!&&\!\!\!\!\!\! + \omega_{\mathbf{d}}||\mathbf{d_k}\! -\!  \mathbf{\bar{d}_k}||^2 \!+\! \omega_{\lambda}||\log(\lambda_k\!+\!1)\! -\!  \log(\bar{\lambda}_{k}\!+\!1)||^2  \Big) \nonumber
\end{eqnarray} 

\begin{table}[t]
\centering
\begin{minipage}[c]{0.5\textwidth}
\centering
\begin{tabular}{|c|c|c|c|c|c|}
\hline
$\omega_{\text{sun}}$ & $\omega_{\text{sky}}$ & $\omega_{\text{grd}}$ & $\omega_{\mathbf{w}}$ & $\omega_{\mathbf{d}}$ & $\omega_{\lambda}$  \\
\hline
1.0 & 0.2 & 0.2 & 0.001 & 1.0 & 0.001 \\
\hline
\end{tabular}
\end{minipage}\hfill
\begin{minipage}[c]{0.5\textwidth}
\vspace{0.3cm}
\caption{Value of coefficients for direct light source loss $\mathbf{Loss_{src}}$.}
\label{tab:sup_srcLoss}
\end{minipage}
\vspace{-0.2cm}
\end{table}

The values of coefficients $\omega_{\cdot}$ are summarized in Tab.~\ref{tab:sup_srcLoss}, which are determined by first fine-tuning on a small training set and then applying to the whole dataset. For invisible windows and lamps, we use the same loss functions as in the visible cases. Note that we only predict one invisible lamp and one invisible window for each image. When there is no invisible window or lamp, we only compute the rendering loss by setting $\mathbf{\bar{E}_j = 0}$. When there are more than one invisible windows or lamps, we pick up the one whose direct shading $\mathbf{\bar{E}_j}$ has the highest total energy and compute all losses with respect to it.

\subsection{Shadow prediction.} 
\begin{figure}[t]
\centering 
\begin{minipage}[c]{0.48\textwidth}
\centering
\includegraphics[width=\columnwidth]{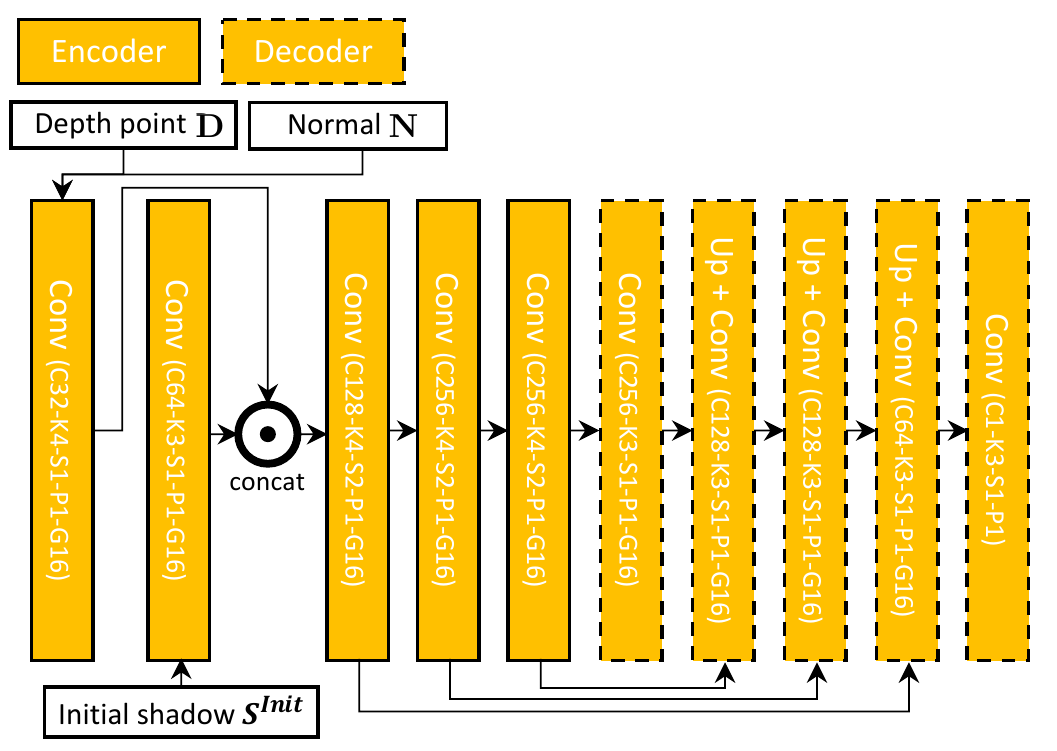}
\vspace{-0.6cm}
\caption{Network architecture for shadow prediction.}
\label{fig:sup_shadow}
\end{minipage}\hfill 
\begin{minipage}[c]{0.50\textwidth}
\centering
\includegraphics[width=\columnwidth]{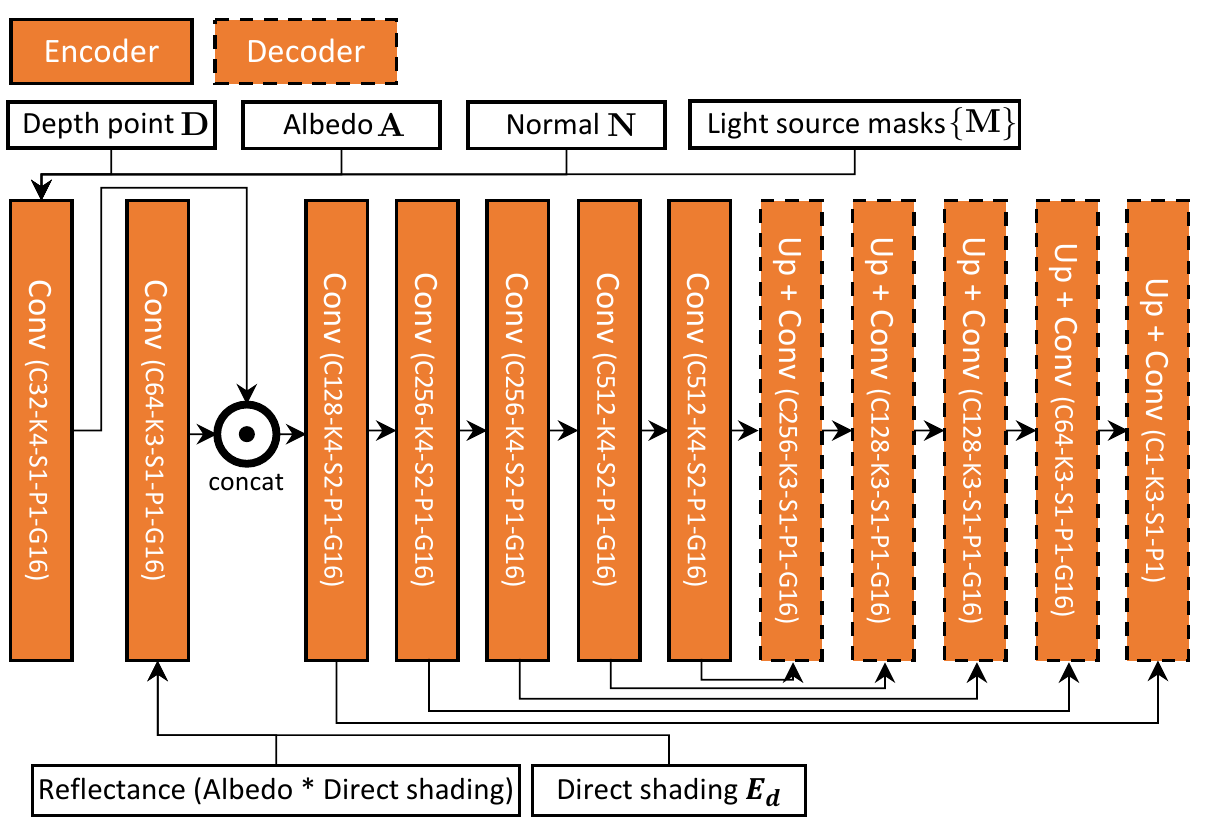}
\vspace{-0.6cm}
\caption{Network architecture for indirect shading prediction.}
\label{fig:sup_indirect}
\end{minipage}
\vspace{-0.2cm}
\end{figure}

\begin{figure}[t]
\centering
\begin{minipage}[c]{0.49\textwidth}
\includegraphics[width=\columnwidth]{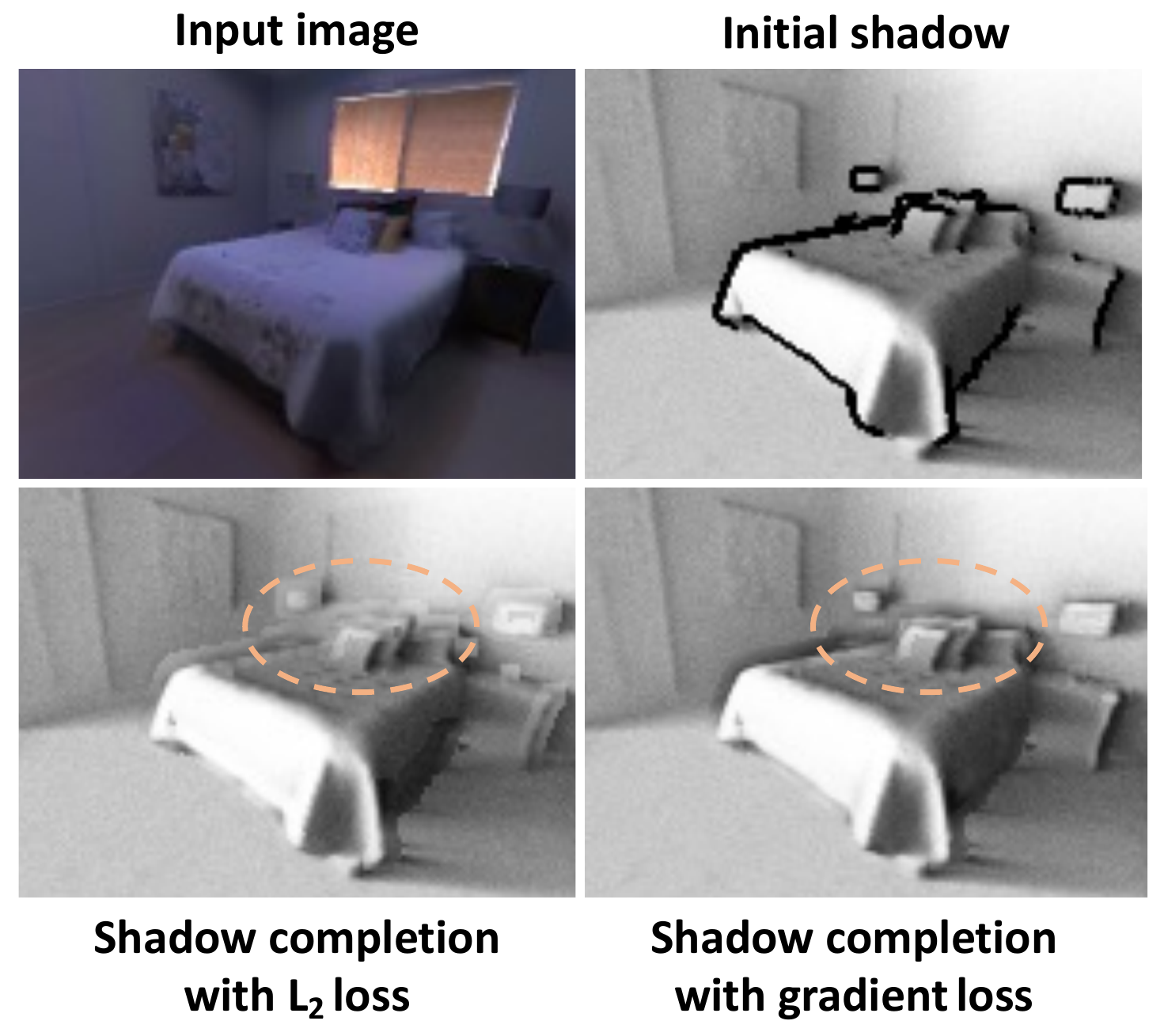}
\end{minipage}\hfill
\begin{minipage}[c]{0.49\textwidth}
\caption{Qualitative comparison of shadow completion network trained with $L_2$ loss and gradient-based loss. The orange circle highlights the artifacts in the result trained with $L_2$ loss. }
\label{fig:sup_shdLoss}
\end{minipage}
\vspace{-0.2cm}
\end{figure}

For shadow prediction, we train $\mathbf{DShdNet}$ to in-paint and denoise the incomplete shadow map due to the occlusion boundaries. The network architecture of  $\mathbf{DShdNet}$ is shown in Fig.~\ref{fig:sup_shadow}, where we use a light-weight encoder-decoder structure with skip-links. We train the shadow prediction network with scale-invariant gradient based loss as proposed in \cite{niklaus20193d} and find that it works much better than standard L2 loss, especially on real images. Formally, let $\mathbf{S}$ and $\mathbf{\bar{S}}$ be the predicted and ground-truth shadows, the loss function is defined as 
\begin{equation*}
\small
\mathbf{Loss_{shd}} = \sum_{h}^{1,2,4,8} \sum_{i, j} ||g_h[\mathbf{S}](i, j) - g_h[\mathbf{\bar{S}}](i, j)||^2,
\end{equation*}
where 
\begin{equation*}
\small
g_h[\mathbf{S}](i, j)\! =\!\! \left(\frac{\mathbf{S}(i\!+\!h, j)\! -\! \mathbf{S}(i, j)}{|\mathbf{S}(i\!+\!h, j)\!+\! \mathbf{S}(i, j)|}, \frac{\mathbf{S}(i, j\!+\!h)\!-\! \mathbf{S}(i, j)}{|\mathbf{S}(i, j\!+\!h)\! + \!\mathbf{S}(i, j)|} \right).
\end{equation*}
Fig.~\ref{fig:sup_shdLoss} compares the shadow completion network trained with standard $L_2$ loss and the gradient-based loss on a real image from the Replica dataset \cite{straub2019replica}. We observe that compared to $L_2$ loss, gradient-based loss leads to smoother prediction with fewer artifacts. 

\subsection{Indirect illumination prediction} 
The network architecture for indirect shading prediction is shown in Fig.~\ref{fig:sup_indirect}. which has an encoder-decoder architecture with a large receptive field so that the network can learn the non-local global information. We train this module independently by taking the ground-truth direct shading $\mathbf{\bar{E}_d}$ as input and predict the indirect shading. 
\begin{eqnarray}
\mathbf{E_{ind}} &=&  \mathbf{IndirectNet(\bar{E}_d, D, N, A)}\\
\mathbf{E} &=& \mathbf{E_{ind} + \bar{E}_d } 
\end{eqnarray}
The loss function is the $L_1$ loss of the final shading prediction $\mathbf{Loss_{shg}}= | \mathbf{E} - \bar{\mathbf{E}}|$.

\subsection{Per-pixel lighting prediction} 
\begin{figure}[t]
\centering
\begin{minipage}[c]{0.6\textwidth}
\includegraphics[width=\columnwidth]{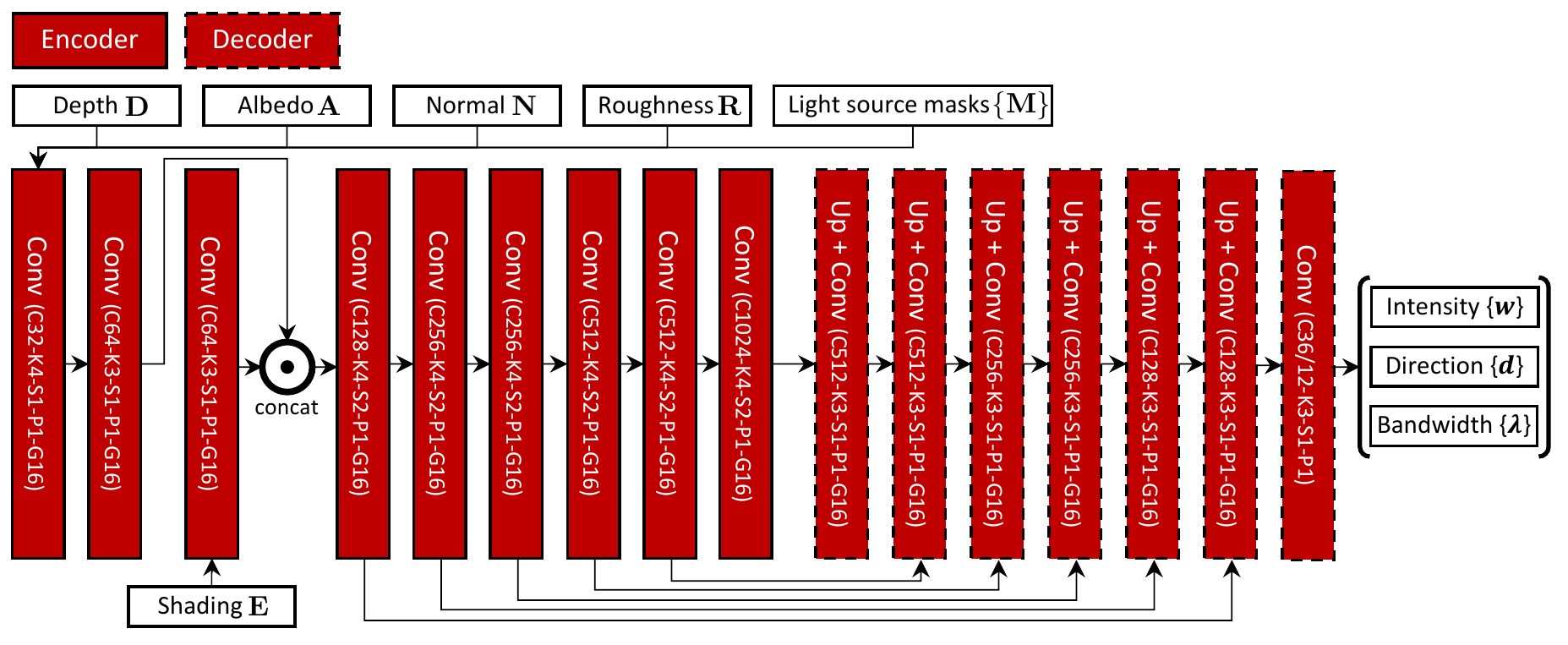}
\end{minipage}\hfill
\begin{minipage}[c]{0.38\textwidth}
\caption{Network architecture for lighting prediction.}
\label{fig:sup_light}
\end{minipage}
\vspace{-0.2cm}
\end{figure}

As shown in Fig.~\ref{fig:sup_light}, the network architecture of $\mathbf{LightNet}$ for per-pixel lighting prediction is exactly the same as in \cite{li2020inverse}. The only difference is that we replace the input image $\mathbf{I}$ with the ground-truth per-pixel shading $\mathbf{\bar{E}}$. 
\begin{equation}
\mathbf{L} = \mathbf{LightNet(\bar{E}, M, A, N, R, D)}
\end{equation}
The loss function is the $\log$ $L_2$ loss on the predicted per-pixel lighting plus a $L_1$ shading loss. To compute the shading loss, we integrate the per-pixel lighting $L$ to get shading prediction $\mathbf{E^L}$ and compute its difference from $\mathbf{\bar{E}}$.  
\begin{equation*}
\small
\mathbf{Loss_{light}} = ||\log(\mathbf{L}+1)\!-\!\log(\mathbf{\bar{L}}+1) ||^2 + \omega_{r} |\mathbf{E}^{L} - \mathbf{\bar{E}}|
\end{equation*}
where $\omega_r$ is set to be 0.01.

\begin{table}[t]
\centering
\small
\begin{tabular}{|c|c|c|c|c|c|c|c|c|}
\hline
&  \multirow{2}{*}{Mat.} & \multicolumn{2}{|c|}{Window} & \multicolumn{2}{|c|}{Lamp} & \multirow{2}{*}{Shadow} & \multirow{2}{*}{Indirect} & \multirow{2}{*}{Lighting} \\
\cline{3-6}
& & Vis. & Inv. & Vis. & Inv. & & & \\
\hline
Iter(k) & 135 & 120 & 200 & 120 & 150 & 70 & 180 & 240 \\
\hline
Batch & 12 & 9 & 9 & 9 & 9 & 3 & 3 & 3 \\
\hline
\end{tabular}
\caption{The number of iterations and batch size for training each network. For all networks, we use Adam optimizer with learning rate $10^{-4}$ and $\mathtt{betas}=(0.9, 0.999)$. }
\label{tab:sup_train}
\vspace{-0.6cm}
\end{table}

\subsection{Differentiable Direct Shading Rendering Layer}
\label{sec:sup_render}
We first discuss how we uniformly sample the surface of light sources. Let $u, v$ be two random variables sampled from a uniform distribution in the range of $[-1, 1]$. To sample a point on the window surface, we have 
\begin{equation}
\mathbf{q} = \mathbf{c} + 0.5u\mathbf{x} + 0.5v\mathbf{y}
\end{equation}
The invisible lamp is represented by a 3D bounding box. Thus, we sample each of the 6 faces separately, for example, 
\begin{equation}
\mathbf{q} = \mathbf{c} + 0.5 \mathbf{x} + 0.5u\mathbf{y} + 0.5v\mathbf{z}. 
\end{equation}
For visible lamps, we treat every visible point $\mathbf{q}$, invisible point $\mathbf{\hat{q}}$ and edge point $\mathbf{q_e}$ as a plate whose normal $\mathbf{N}$ and area can be computed as in Eq.~\eqref{eq:sup_areaVis}-\eqref{eq:sup_nEdge}. Replacing these sampled points into Eq.~\ref{eq:areaSample} in the main paper allows us to compute the direct shading for every light source. 

For window light sources specifically, we observe that it is necessary to sample according to the angular distribution of high-frequency directional sunlight (Fig.~\ref{fig:winRep} in the main paper). To achieve this, we use standard Monte Carlo sampling by first computing the CDF of $\mathcal{G}_{\text{sun}} = \{\mathbf{w}_{\text{sun}}, \mathbf{d}_{\text{sun}}, \lambda_{\text{sun}}\}$ and then sampling the lighting direction using its inverse function. We define $\theta_{\text{sun}}$ and $\phi_{\text{sun}}$ to be the polar angle and azimuthal angle in a coordinate system where $\mathbf{d}_\text{sun}$ is the $\mathbf{z}$ axis. The PDF for $\theta_{\text{sun} }$ and $\phi_{\text{sun}}$ are
\begin{eqnarray*}
\small
\mathbf{Pr}(\phi_{\text{sun}}) &=& \frac{1}{2\pi}\\
\mathbf{Pr}(\theta_{\text{sun}}) &=& \frac{\lambda_{\text{sun}}\exp(\lambda_{\text{sun}}(\cos \theta_{\text{sun}} - 1 )\sin\theta_{\text{sun}}}{1 - \exp(-2\lambda_{\text{sun}}) }   \\
\mathbf{Pr}(\phi_{\text{sun}}, \theta_{\text{sun}}) &=& \mathbf{Pr}(\phi_{\text{sun}})\mathbf{Pr}(\theta_{\text{sun}})
\end{eqnarray*}
Similar as when we sample the surface of light sources, let $u, v$ be two random variables. To sample $\phi_{\text{sun}}$, since it distributes uniformly, we simply have 
\begin{equation*}
\small
\phi_{\text{sun}} = u\pi
\end{equation*}
To sample $\theta_{\text{sun}}$, we first compute the CDF of its distribution
\begin{equation*}
\small
\mathbf{F}(\theta_{\text{sun}}) = \frac{\exp(\lambda_{\text{sun}}) - \exp(\lambda_{\text{sun}}\cos\theta_{\text{sun}}) }{\exp(\lambda_{\text{sun}}) - \exp(-\lambda_{\text{sun}}) },
\end{equation*}
and then compute $\theta_{\text{sun}}$ by inverting its CDF 
\begin{equation*}
\small
\theta_{\text{sun}} = \arccos\left(\frac{\log(1 - \frac{v+1}{2}(1 - \exp(-2\lambda_{\text{sun}})))}{\lambda_{\text{sun}}} + 1 \right)
\end{equation*}
Given $\theta_{\text{sun}}$ and $\phi_{\text{sun}}$, the sampled lighting direction $\mathbf{l}$ can be computed as
\begin{eqnarray*}
\small
\mathbf{l} &=& \mathbf{x}_{\text{sun}}\sin\theta_{\text{sun}}\cos\phi_{\text{sun}} + \mathbf{y}_{\text{sun}}\sin \theta_{\text{sun}}\cos\phi_{\text{sun} } \\
&& + ~~\mathbf{d}_{\text{sun}}\cos\theta_\text{sun} 
\end{eqnarray*}
which can be replaced into Eq.~\ref{eq:angSample} in the main paper to compute the direct shading. Here $\mathbf{x}_{\text{sun}}$ and $\mathbf{y}_{\text{sun}}$ are two arbitrary orthogonal vectors perpendicular to $\mathbf{d}_{\text{sun}}$. Note that we implement all the sampling algorithms in pytorch so that our rendering layer is differentiable. 

\begin{figure}[t]
\centering
\begin{minipage}[c]{0.49\textwidth}
\includegraphics[width=\columnwidth]{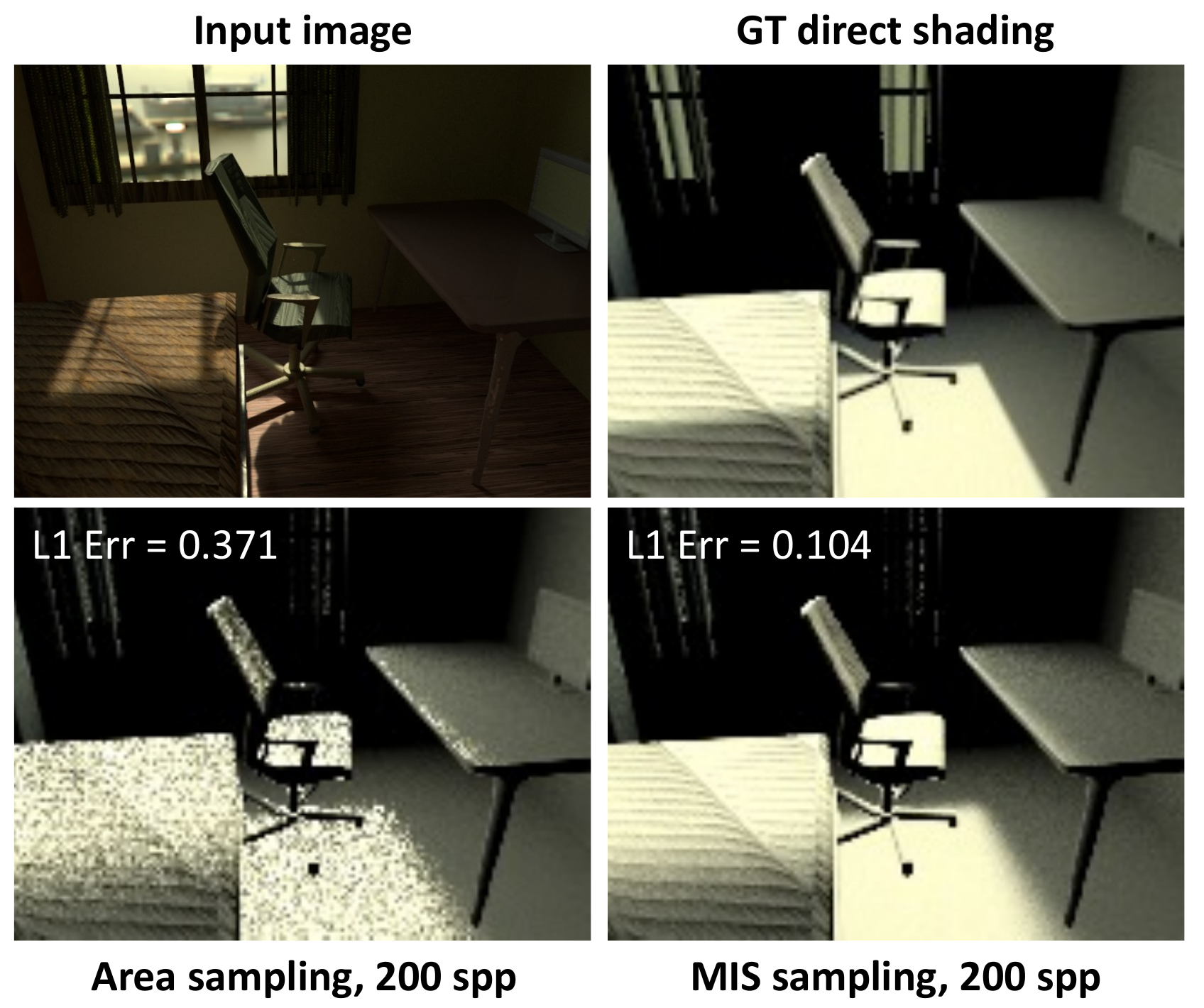}
\end{minipage}\hfill
\begin{minipage}[c]{0.49\textwidth}
\caption{Comparions of direct shading rendered by sampling area uniformly or using MIS sampling. Our MIS sampling has much less noise with the same number of samples. This makes it possible for us to train our networks with rendering loss, which is essential to achieve accurate light source reconstruction. }
\label{fig:sup_mis}
\end{minipage}
\vspace{-0.2cm}
\end{figure}

\begin{figure}[t]
\centering
\begin{minipage}[c]{0.6\textwidth}
\includegraphics[width=\columnwidth]{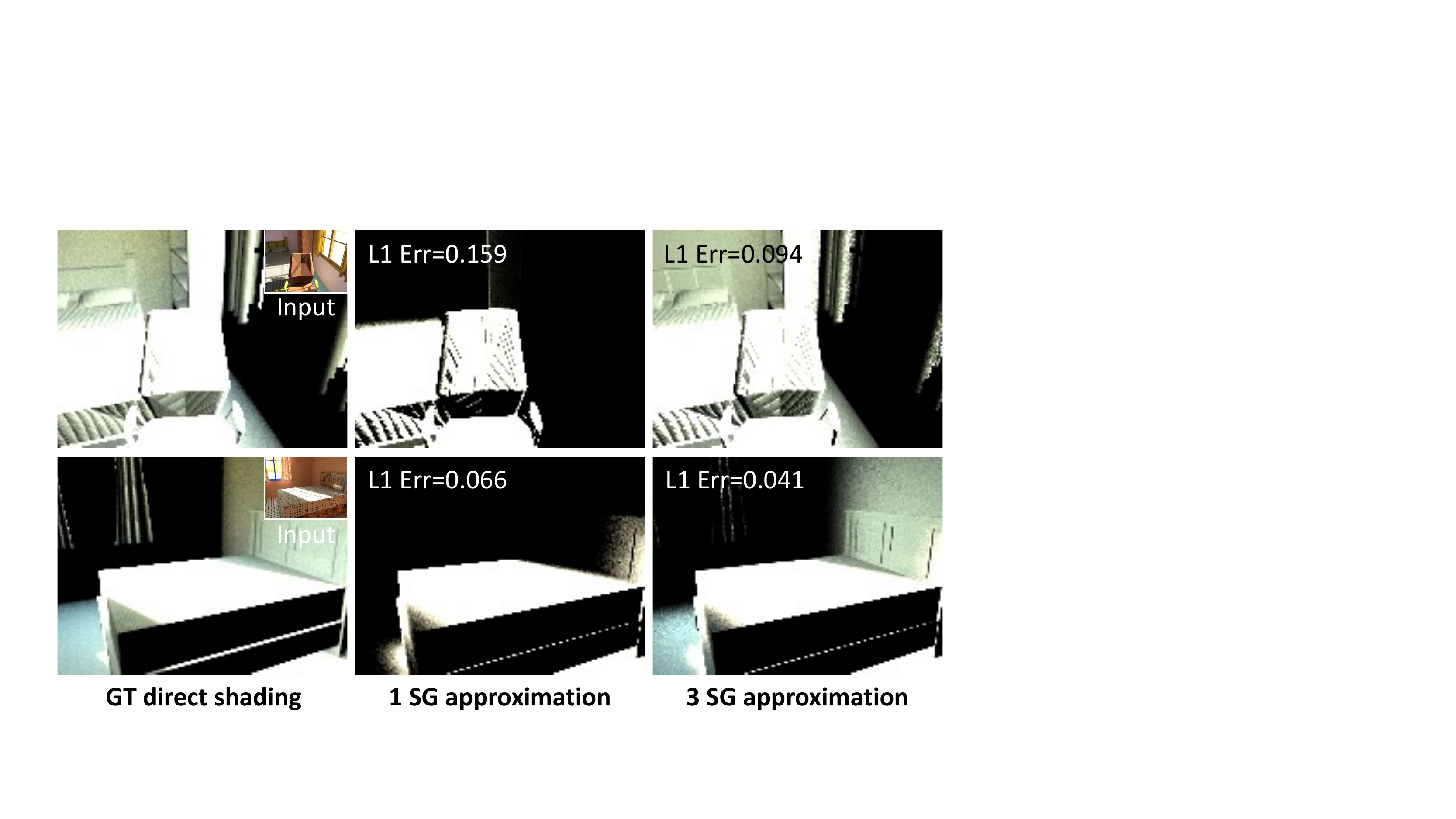}
\end{minipage}\hfill
\begin{minipage}[c]{0.38\textwidth}
\caption{A demonstration of direct shading rendered from our ground-truth window radiance parameters. Our ground-truth 3 SGs can be used to render direct shading that closely matches the ground-truth and is more expressive compared to a single SG representation, which cannot capture the ambient lighting.}
\label{fig:sup_gtWin}
\end{minipage}
\vspace{-0.2cm}
\end{figure}

\begin{figure*}[t]
\centering
\includegraphics[width=\textwidth]{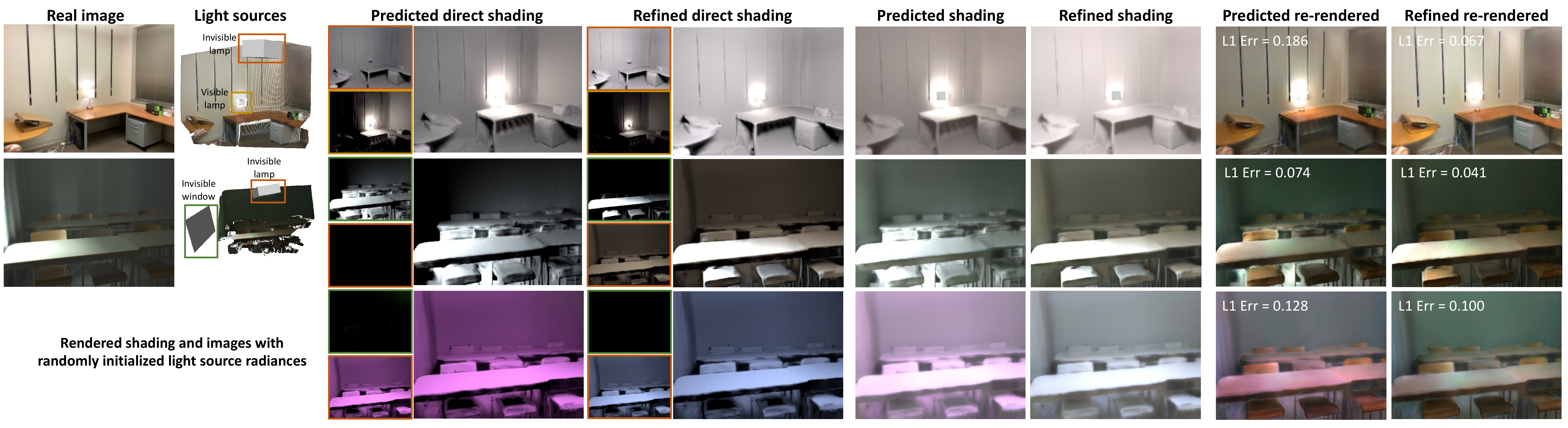}
\vspace{-0.6cm}
\caption{Comparisons of light source prediction and re-rendered image before and after optimization. We observe that while the optimization-based refinement can help predict more consistent light source intensity, it also relies on a good initialization from the network to converge to a good result: a random initialization cannot lead to accurate recovery of light source radiance through pure optimization. This is especially true for more complex sunlight coming through windows. Note that the direct shading from the invisible window for the first example is always $\mathbf{0}$ and therefore is not shown here. }
\label{fig:sup_optim}
\vspace{-0.2cm}
\end{figure*}

\begin{table}[t]
\centering
\small
\begin{tabular}{|c|c|c|c|c|c|c|}
\hline
\multirow{2}{*}{Material} & Light & Direct & \multirow{2}{*}{Shadow} & Indirect & Per-pixel & \multirow{2}{*}{Total} \\
& source & shading & & shading & lighting & \\
\hline
299ms  & 19.7ms  & 595ms   & 1309ms  & 19.1ms  &19.35ms & 2.26s  \\
\hline
\end{tabular}
\caption{Inference time of each step of our framework. }
\vspace{-0.6cm}
\label{tab:sup_time}
\end{table}

In Fig.~\ref{fig:winRep} in the main paper and Fig.~\ref{fig:sup_mis}, we demonstrate that sampling according to both the geometry and radiance distribution of a window following the MIS rule can lead to much less noise with similar number of samples, compared to only uniformly sampling the surface area of the window.

\subsection{Ground-truth Window Radiance Parameters}
\label{sec:sup_gtLightSrc}
Our Monte Carlo-based differentiable direct shading rendering layer allows us to compute ground-truth radiance parameters for windows, by minimizing the rendering loss between the rendered direct shading $\mathbf{E}_{\mathcal{W}}$ and the ground-truth direct shading $\mathbf{\bar{E}}_{\mathcal{W}}$ provided by the OpenRooms dataset, 
\begin{equation}
\small
\mathbf{Loss}_{\mathbf{ren}} = |\mathbf{E}_{\mathcal{W}} - \mathbf{\bar{E}}_{\mathcal{W}}| 
\end{equation}
Here we use $L_1$ loss instead of $\log L_2$ loss because we observe that the latter can recover low-intensity regions more accurately but meanwhile can lead to highlight artifacts. To encourage the 3 SGs to represent 3 physically meaningful light sources, sun, sky and ground respectively, we first render a panorama facing outside the window and then select the brightest direction in the panorama as the sunlight direction and keep it fixed through the optimization process. As for the other 2 SGs corresponding to sky and ground, we initialize their direction with up vector [0, 1, 0] and minus up vector [0, -1, 0] in the world coordinate system. In addition, we also apply the $\lambda$ constraint as shown in Tab.~\ref{tab:sup_lambda} so that the high-frequency directional lighting can be mainly represented by the $\mathcal{G}_{\text{sun}}$. 

In Fig.~\ref{fig:winRep} in the main paper and Fig.~\ref{fig:sup_gtWin}, we demonstrate that our ground-truth 3 SGs parameters can be used to render direct shading very close to the ground-truth, with both high-frequency directional lighting and ambient lighting being correctly modeled, while the 1 SG representation applied by prior work \cite{wang2021learning} can only capture the directional lighting. Our ground-truth 3 SGs parameters are used to compute the $\mathbf{Loss_{src}^{\mathcal{W}}}$ as shown in Eq.~\eqref{eq:sup_srcLoss} in the training process and demonstrated to help capture more accurate and interpretable lighting in Sec.~\ref{sec:sup_exp}. 

\subsection{Optimization-based Refinement}
\label{sec:sup_optim}

Our differentiable rendering pipeline allows us to refine the light source radiance parameters based on rendering loss. We find that this is especially useful when the intensity of light source prediction can be slightly off sometimes. Given the light source parameters $\{\mathcal{W}\}$ and $\{\mathcal{L}\}$, which cover visible/invisible and windows/lamps, we can render shading $\mathbf{E}$. We define the rendering loss as the $L_2$ loss between the rendered image and the input LDR image, 
\begin{equation}
\mathbf{Loss_{ren}} = ||\min(\mathbf{EA}, 1) - \mathbf{I}||^2,
\end{equation}
where $\mathbf{A}$ is the predicted albedo. Note that we have already transformed the input LDR image into linear RGB space. One alternative to compute the rendering loss is to use per-pixel lighting $\mathbf{L}$ so that we can also render specularity. However, we observe that it will cause the optimization to be unstable. 

Fig.~8 and Fig.~\ref{fig:sup_optim} compares the light source prediction and re-rendered image before and after optimization, where we observe that our rendering error-based optimiztion can effectively correct the intensity of the light source prediction. However, we also observe that for more complex sunlight coming through a window, it is important to provide a good initial prediction from the network. Otherwise, a pure optimization-based method cannot recover light source radiance correctly. In the second example in Fig.~\ref{fig:sup_optim}, we randomly initialize the light source radiance and observe that reconstructed direct shading and final re-rendered image may not be accurate.

\subsection{Inference Time}
The inference time for each step of the network to process one image is summarized in Tab.~\ref{tab:sup_time}. The most time consuming step is to render shadows from depth using path tracing. Note that while our framework handles many complex light transport effects, including global illumination, the total time for it to reconstruct and re-render an indoor scene is only less than 3 s.

\section{Synthetic Experiments on OpenRooms}
\label{sec:sup_exp}

\begin{figure*}[t]
\centering
\includegraphics[width=\textwidth]{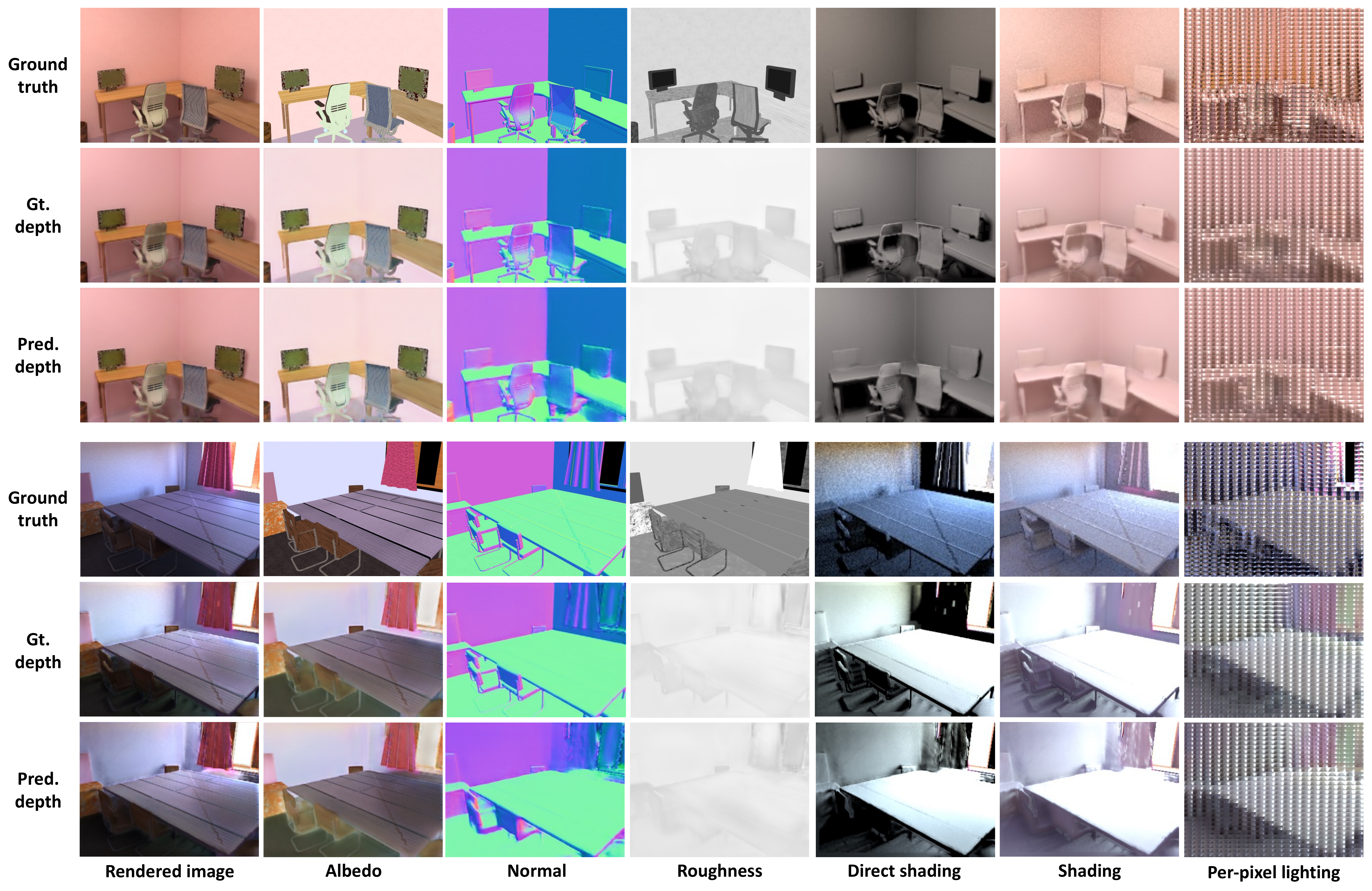}
\vspace{-0.6cm}
\caption{Material predictions and neural rendering results on the OpenRooms synthetic dataset with predicted and ground-truth depth.  We observe that with both ground truth  our method can render high-quality direct shading, shading, per-pixel environment map and final image from our light source and material predictions, with non-local shadows and interreflection being correctly modeled. }
\vspace{-0.2cm}
\label{fig:sup_synpred}
\end{figure*}

\begin{figure*}[t]
\centering
\includegraphics[width=\textwidth]{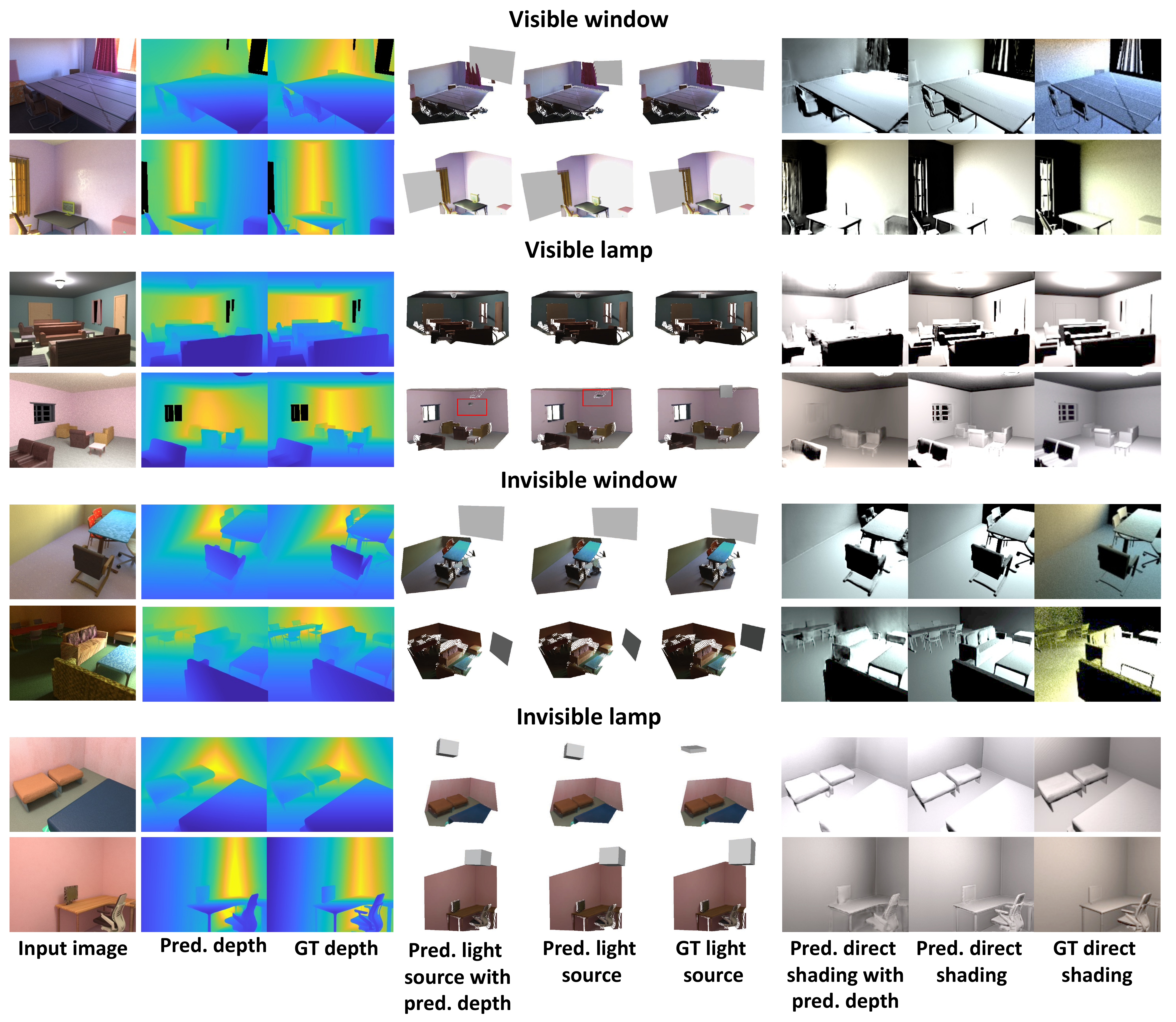}
\vspace{-0.6cm}
\caption{Light source prediction results on the synthetic dataset for various types of light sources with ground-truth and predicted depth. In most cases, our method can recover both the geometry and radiance of light sources similar to ground truth with either predicted or ground-truth depth. We also show one example on the fourth row, as marked by the red rectangle, where the inaccurate depth prediction leads to poor geometry prediction of a visible lamp, causing the highlight in the shading to be missing. }
\label{fig:sup_synsrc}
\vspace{-0.2cm}
\end{figure*}

\begin{figure}[t]
\centering
\begin{minipage}[c]{0.49\textwidth}
\includegraphics[width=\columnwidth]{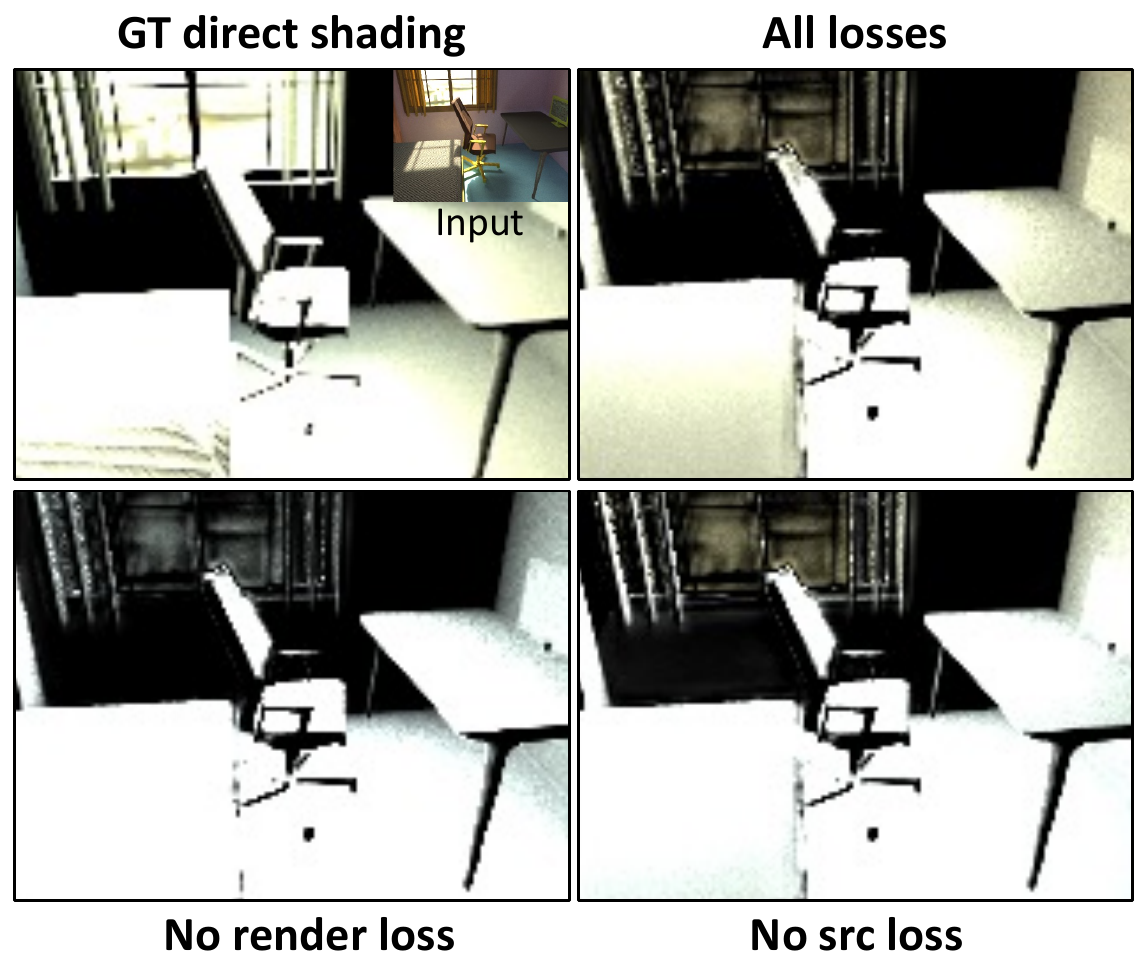}
\end{minipage}\hfill
\begin{minipage}[c]{0.49\textwidth}
\caption{Ablation studies on different loss combinations for window light source prediction. Our network trained with both rendering loss $\mathbf{Loss_{ren}^{j}}$ and light source loss $\mathbf{Loss_{src}^{j}}$ predicts the most accurate radiance, with both high-frequency directional lighting and ambient lighting closely matching the ground-truth. }
\label{fig:synablation}
\end{minipage}
\vspace{-0.2cm}
\end{figure}

\begin{table}[t]
\centering
\small
\setlength{\tabcolsep}{2pt}
\begin{tabular}{|c|c|c|c|c|c|c|}
\hline
& \multicolumn{2}{|c}{Albedo $10^{-2}$} & \multicolumn{2}{|c}{Normal $10^{-2}$} & \multicolumn{2}{|c|}{Roughness $10^{-2}$}  \\
& \multicolumn{2}{|c}{$\mathbf{A}$} & \multicolumn{2}{|c}{$\mathbf{N}$} & \multicolumn{2}{|c|}{$\mathbf{R}$} \\ 
\hline
\multirow{2}{*}{Ours} & Gt. & Pred. & Gt. & Pred. & Gt. & Pred.  \\ 
\cline{2-7} 
&1.81 & 2.48 & 1.39 & 6.52 & 6.22 & 6.58  \\
\hline
Li et al. \cite{li2020openrooms} & \multicolumn{2}{|c}{-} & \multicolumn{2}{|c}{4.51} & \multicolumn{2}{|c|}{6.59} \\
\hline
\end{tabular}
\normalsize
\caption{Material predictions on the OpenRooms testing set. We report $L_2$ error of our material predictions. We report our results with both ground-truth depth and predicted depth as inputs. The network is trained with ground-truth depth and not fine-tuned with predicted depth.}
\vspace{-0.6cm}
\label{tab:sup_mat}
\end{table}

\begin{figure}[t]
\centering
\includegraphics[width=0.8\textwidth]{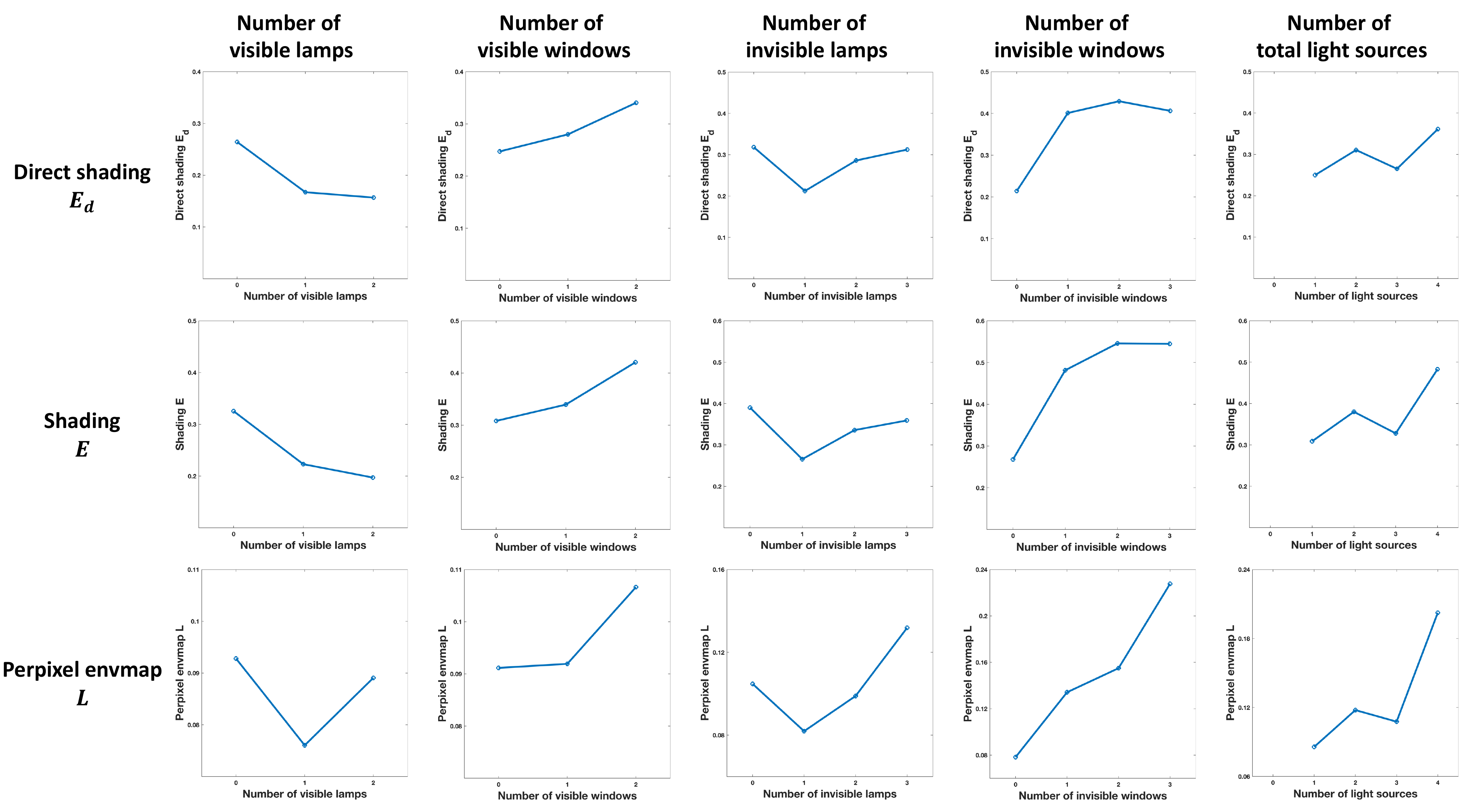}
\vspace{-0.6cm}
\caption{Rendering error distribution with respect to the number of light sources. We observe that error increases when the number of windows increases, which is because the radiance of windows are more complex and difficult to predict.}
\label{fig:sup_errorDist}
\vspace{-0.2cm}
\end{figure}

We present more qualitative and quantitative results on the synthetic OpenRooms dataset \cite{li2020openrooms}. More specifically, we test the effectiveness of different loss terms and how imperfect depth prediction can impact our light source prediction and neural rendering results. Our depth prediction are produced by DPT \cite{ranftl2021vision} without fine-tuning on our synthetic dataset. We train all our models on ground-truth depth and, as shown in multiple figures (e.g., Fig. \ref{fig:teaser}, \ref{fig:garonPlus} and \ref{fig:onOff}) in the main paper and Fig. \ref{fig:sup_synpred} and \ref{fig:sup_synsrc}, find that they generalize well to predicted depth for both real and synthetic data in most cases.

\subsection{Material prediction} 

Unlike the prior method \cite{li2020inverse}, which first uses scale-invariant loss for albedo prediction and adopts a linear regression to solve the scale ambiguity, we use the absolute loss for both diffuse albedo and light intensity prediction. The reason is that our method needs to recover the radiance of multiple light sources in the scene and it is difficult to recover consistent intensities across multiple light sources through simple linear regression. 

Tab.~\ref{tab:sup_mat} compares our material prediction with \cite{li2020openrooms}. We report the quantitative numbers with both ground-truth and predicted depth maps as inputs. When using ground-truth depth as an input, our normal prediction is much more accurate compared to \cite{li2020openrooms}.  Our roughness quality is similar to \cite{li2020inverse}. Both the roughness and albedo predictions are relatively insensitive to the depth accuracy. In Fig.~\ref{fig:sup_synpred} and Fig.~\ref{fig:prediction} in the main paper, we present our material predictions on both real and synthetic data. On synthetic data, we show that our diffuse albedo, roughness and normal predictions are reasonably close to the ground truths. For real images, even though we do not have ground-truths, our material predictions are high-quality enough to enable realistic re-rendering of the scene.

\subsection{Light source prediction}
In Fig.~\ref{fig:sup_synsrc}, we show more qualitative light source prediction results using either ground-truth or predicted depth. The quantitative numbers are summarized in Tab. \ref{tab:src} in the main paper. In most cases, our light source prediction models, even though trained on ground-truth depth only, can generalize well to predicted depth and can recover geometry and radiance of all 4 types of light sources accurately. In the reconstructed direct shading, small errors can be seen caused by the imperfect depth prediction with less details, which might be inevitable. Our visible lamp reconstruction is more sensible to depth accuracy compared to other kinds of light sources due to its geometry representation. In the fourth row, we show one example where the inaccurate depth prediction causes the lamp position to be closer to the camera than the ground truth. Hence, the highlights on the floor is missing. This example may suggest that utilizing lighting information to improve geometry reconstruction can be an interesting future direction. 

\vspace{-0.3cm}
\paragraph{Ablation study}
Tab.~\ref{tab:sup_ablation} and Fig.~\ref{fig:synablation} verifies the effectiveness of our loss functions for window light source prediction. We observe that while training with light source loss $\mathbf{Loss_{src}^{j}}$ can lead to the prediction closest to our optimized ground-truth light source parameters, the rendering error is significantly higher because it is difficult to find the best balance across different parameters that can minimize the rendering error.  Training with $\mathbf{Loss_{ren}^{j}}$ alone leads to reasonable direct shading prediction. However, the light source parameters are less interpretable, as shown in Tab.~\ref{tab:sup_ablation} and the rendered direct shading tends to be oversmoothed, as shown in Fig~\ref{fig:synablation}.  Combining the two losses together, on the contrary, allows us to render direct shading closer to the ground-truth, with high-frequency lighting being correctly modeled, as shown in both Tab.~\ref{tab:sup_ablation} and Fig.~\ref{fig:synablation}. 

\subsection{Neural rendering}
In Fig.~\ref{fig:sup_synpred}, we also show more neural rendering results with both predicted and ground-truth depth. Our physically-based neural rendering module is reasonably robust to depth inaccuracy, which can reconstruct high-quality direct shading, shading and per-pixel lighting similar to the ground truths. 

\vspace{-0.3cm}
\paragraph{Error distribution} We report distribution of errors in Tab. \ref{tab:synrender} in the main paper with respect to the number of light sources in Fig. \ref{fig:sup_errorDist}. Error increases when a scene has more windows or total number of light sources. It decreases or fluctuates with more lamps possibly because radiance of lamps can be predicted more accurately.

\begin{table}[t]
\centering
\small
\begin{tabular}{|c|c|c|c|c|}
\hline
Visible & Rendering & \multicolumn{3}{|c|}{Light source}  \\ 
\cline{2-5}
 window & Direct  & Intensity   & Direction & Bandwidth  \\&$\mathbf{E_j}$ & $\mathbf{w}$ & $\mathbf{d}$ & $\lambda$ \\
\hline
 w/o $\mathbf{Loss_{ren}^{j}}$ & 1.276 & \textbf{7.972} & 0.386 & \textbf{4.369} \\ 
\hline
w/o $\mathbf{Loss_{src}^{j}}$ & 0.859 & 17.73 & 0.503 & 7.492 \\
\hline
 All & \textbf{0.849} & 10.28 & \textbf{0.369} & 4.419  \\ 
\hline
Invisible & Rendering & \multicolumn{3}{|c|}{Light source}  \\ 
\cline{2-5}
 window & Direct  & Intensity   & Direction & Bandwidth  \\&$\mathbf{E_j}$ & $\mathbf{w}$ & $\mathbf{d}$ & $\lambda$ \\
\hline
 w/o $\mathbf{Loss_{ren}^{j}}$ & 1.786 & \textbf{10.817} & 0.545 & \textbf{4.770} \\ 
\hline
w/o $\mathbf{Loss_{src}^{j}}$ & 0.334 & 44.04 & 1.432 & 70.48 \\
\hline
 All & \textbf{0.312}  & 18.15  & \textbf{0.536}  & 8.168  \\ 
\hline
\end{tabular}
\caption{Ablation studies on window light source prediction. We report $L_1$ loss for direct shading $E_{j}$, $L_2$ loss for direction $\mathbf{d}$ and $\log L_2$ loss for intensity $\mathbf{w}$ and $\lambda$. }
\vspace{-0.6cm}
\label{tab:sup_ablation}
\end{table}

\begin{figure}[t]
\centering
\begin{minipage}[c]{0.49\textwidth}
\includegraphics[width=\columnwidth]{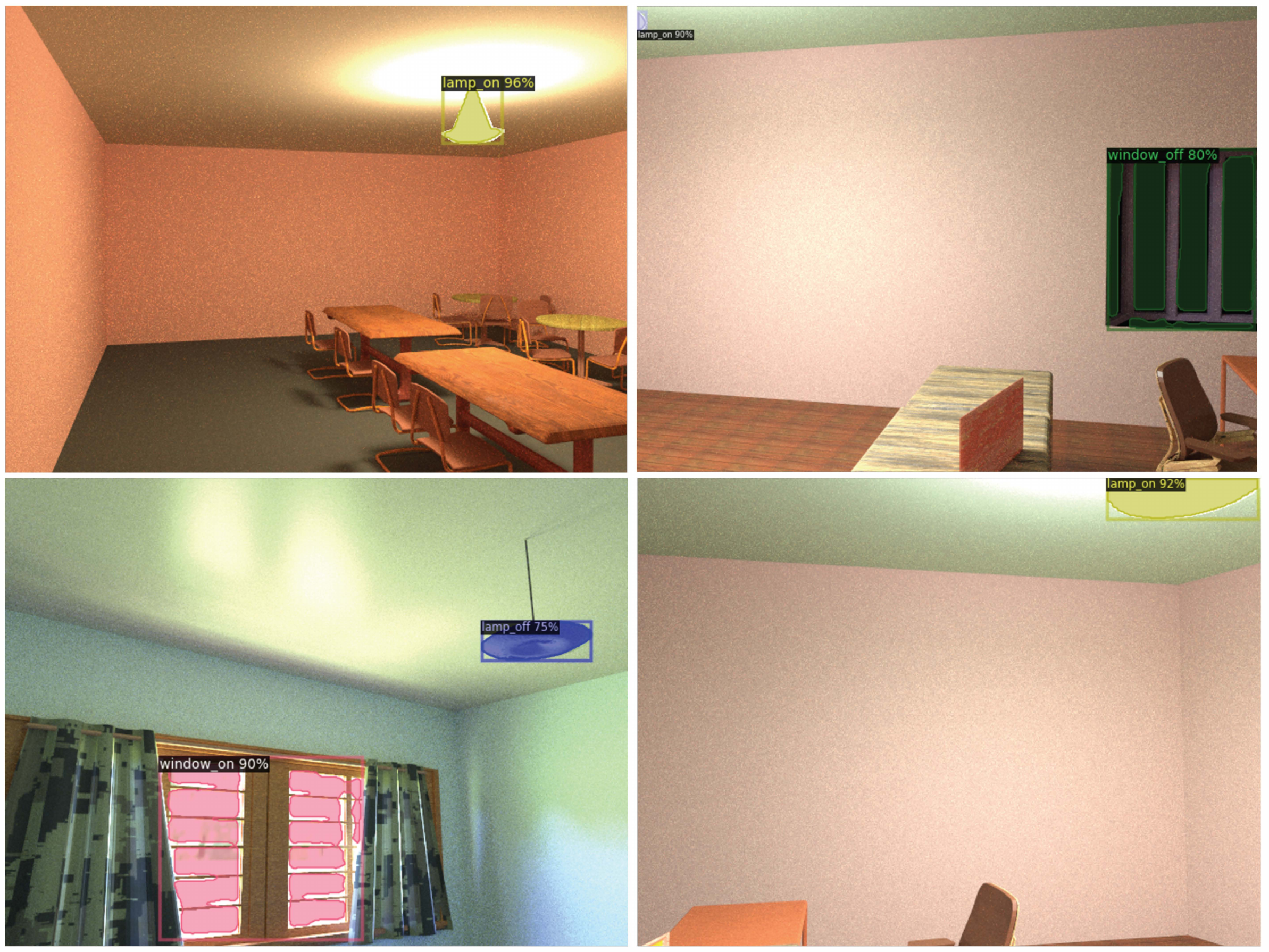}
\end{minipage}\hfill
\begin{minipage}[c]{0.49\textwidth}
\caption{Light source detection and instance segmentation results on the OpenRooms dataset \cite{li2020openrooms}. }
\label{fig:instance}
\end{minipage}
\vspace{-0.2cm}
\end{figure}

\begin{figure}[t]
\centering 
\begin{minipage}[c]{0.6\textwidth}
\includegraphics[width=\columnwidth]{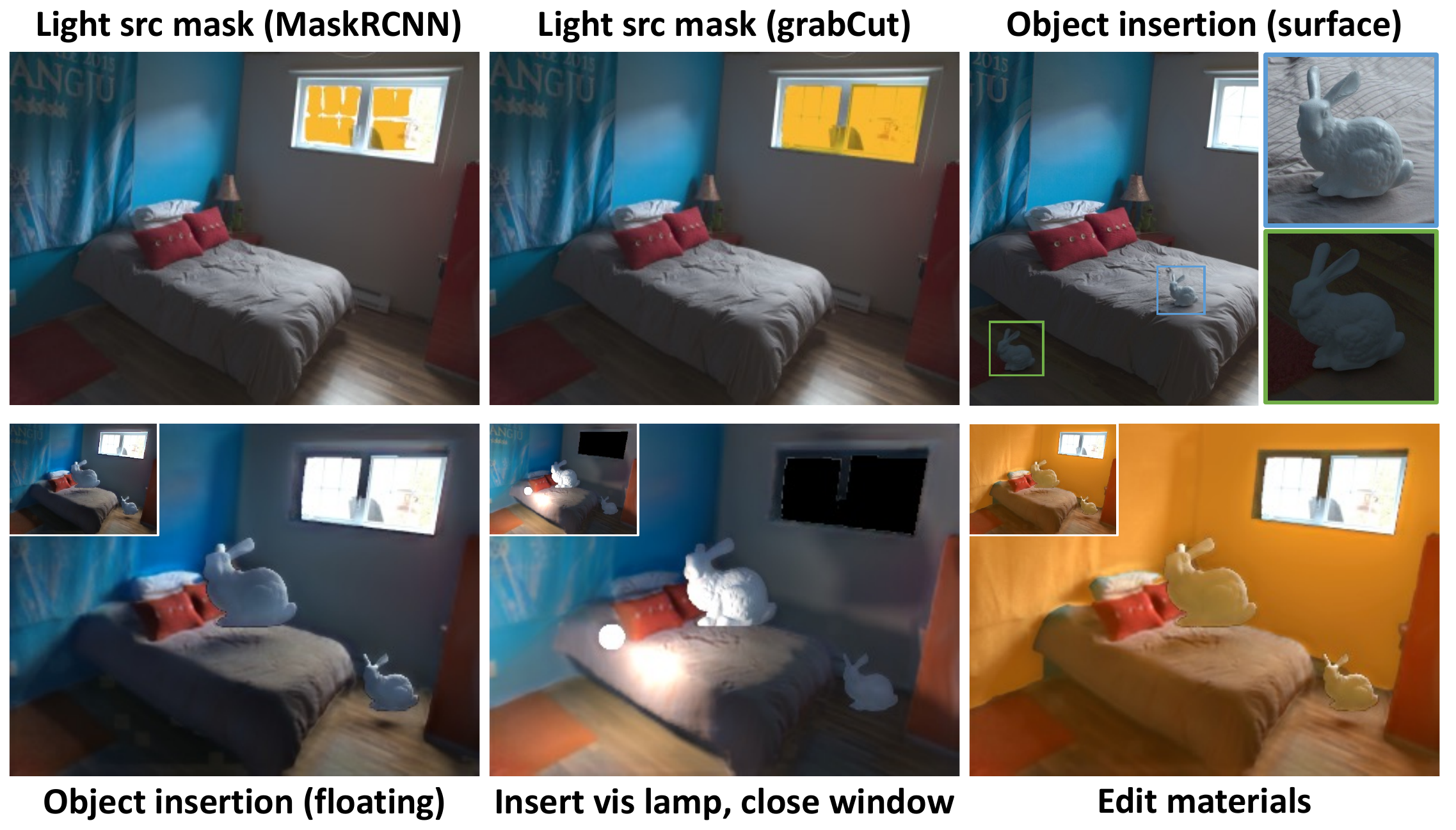}
\end{minipage}\hfill
\begin{minipage}[c]{0.38\textwidth}
\caption{Real image light editing results with predicted light source segmentation mask. The light editing results from a manually created mask are shown in the insets. }
\label{fig:sup_realMask}
\end{minipage}
\vspace{-0.2cm}
\end{figure}

\begin{table}[t]
\small
\begin{minipage}[c]{0.49\textwidth}
\centering
\begin{tabular}{ |c|cc| }
 \hline
  Metric/Type  &  bbox &  seg \\

   \hline
    AP(0.5:0.95)&  65.4 &  59.4 \\
    AR(0.5:0.95)&  85.1 &  78.1 \\
 \hline
    AP-windows-on&  75.4 &  57.0   \\
    AP-lamp-on&  70.4 &  72.1  \\
    AP-windows-off&  54.0 &  50.0  \\
     AP-lamp-off&  61.8 &  63.6  \\
 \hline
\end{tabular}
\end{minipage}\hfill
\begin{minipage}[c]{0.49\textwidth}
\caption{Quantitative evaluation on bounding box regression and mask on OpenRooms\cite{li2020openrooms} for light source (windows and lamps) detection and instance segmentation. }
\label{tab:instance}
\end{minipage}
\vspace{-0.2cm}
\end{table}

\begin{figure}[th]
\centering
\begin{minipage}[c]{0.4\textwidth}
\centering
\includegraphics[width=\columnwidth]{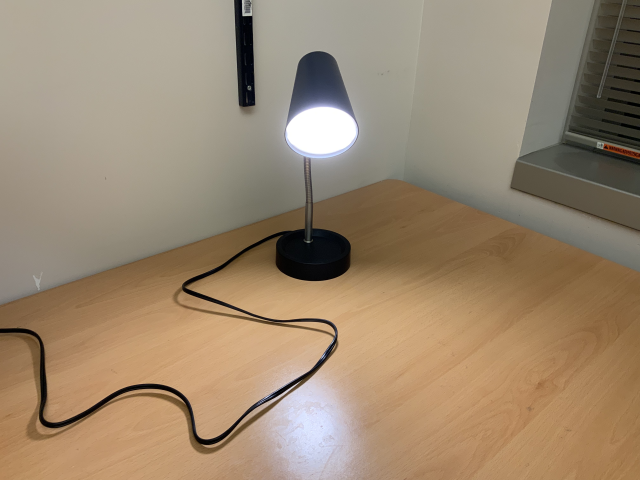}
\end{minipage}\hfill
\begin{minipage}[c]{0.58\textwidth}
\caption{A example where an indoor lamp is not symmetric and does not emit light uniformly in every direction. }
\label{fig:sup_lampFailed}
\end{minipage}
\vspace{-0.2cm}
\end{figure}

\begin{figure}[th]
\centering
\begin{minipage}[c]{0.6\textwidth}
\includegraphics[width=\columnwidth]{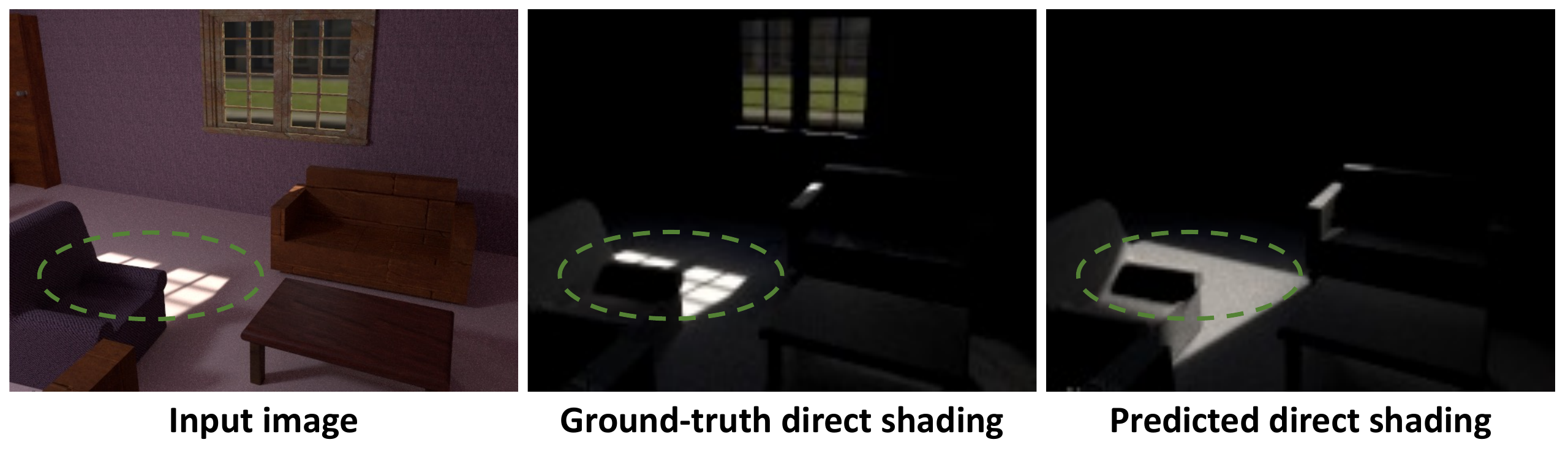}
\end{minipage}\hfill 
\begin{minipage}[c]{0.38\textwidth}
\vspace{-0.3cm}
\caption{While our separation of shading and shadow is necessary for light editing, it can cause missing details in direct shading $\mathbf{E_d}$, as shown in the green circle. }
\label{fig:sup_shgFailed}
\end{minipage}
\vspace{-0.2cm}
\end{figure}

\begin{figure}[t]
\centering
\begin{minipage}[c]{0.49\textwidth}
\centering
\includegraphics[width=\columnwidth]{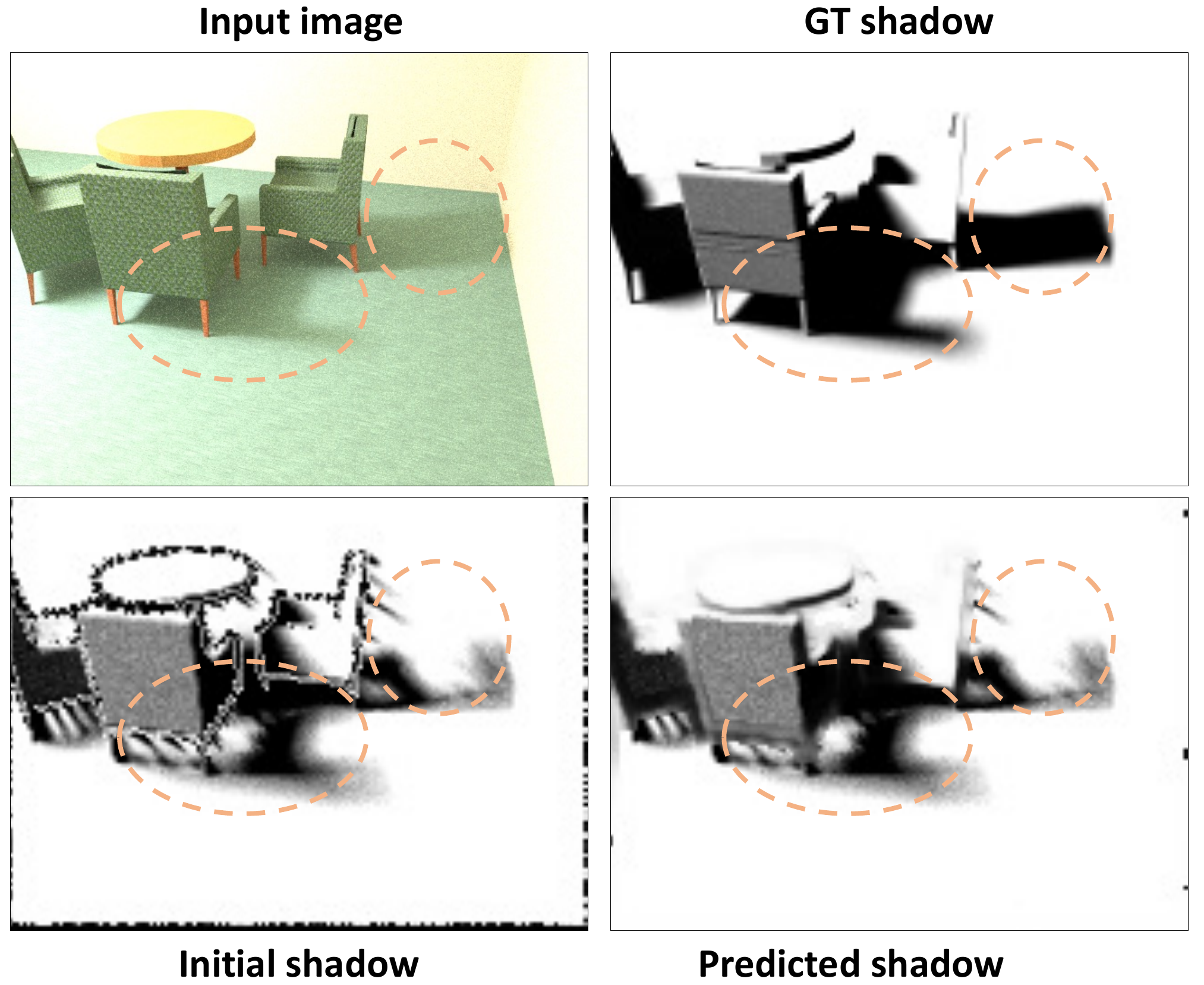}
\end{minipage}\hfill
\begin{minipage}[c]{0.49\textwidth}
\caption{Our shadow rendering framework cannot handle the situation when a large part of the object is not reconstructed due to occlusions, as shown in the errors in the orange circles. }
\label{fig:sup_shdFailed}
\end{minipage}
\vspace{-0.2cm}
\end{figure}

\section{Light Editing with Predicted Masks}
\label{sec:sup_lightEdit}

In all our prior synthetic and real experiments, we assume that the light source segmentation masks are given. While not being our focus, we fine-tuned a Mask RCNN \cite{he2017mask} on the OpenRooms dataset and report its performances.  The fine-tuned Mask RCNN  can detect and segment 4 types of objects, windows and lamps, on and off.  Quantitative and qualitative results are summarized in Tab.~\ref{tab:instance} and Fig.~\ref{fig:instance} respectively, where we observe our fine-tuned model works well on the synthetic dataset. This fine-tuned model can be used for real image editing by providing the initial light source segmentation masks, as will be discussed next.

We apply the fine-tuned MaskRCNN described above on a real image and see if an imperfect light source segmentation mask can still enable high-quality light editing applications. We first use our fine-tuned MaskRCNN to get an initial segmentation mask and then use the GrabCut method \cite{rother2004grabcut} to refine its boundaries. The results are summarized in Fig.~\ref{fig:sup_realMask}. We observe that even though the mask prediction is not perfect, our light editing results are very similar to those shown in Fig.~\ref{fig:garonPlus} in the main paper with a manually created mask, which suggests that our light source prediction and neural renderer can be robust to small mask prediction errors. 

\section{Limitations and Future Works}
\label{sec:sup_limitation}

In this section, we analyze the limitations of our indoor light editing framework. We mainly focus on failure cases caused by our deliberate design choices to highlight the trade-offs being made to build our framework.

\vspace{-0.3cm}
\paragraph{Non-symmetric lamps} Our visible lamp representation assumes that lamps are symmetric with respect to their centers. While this simple assumption holds in many cases, especially for ceiling lamps, it can fail and cause highlight artifacts.  Fig.~\ref{fig:sup_lampFailed} shows an example where the geometry of the lamp cannot be simply represented by reflecting its visible area. Our visible lamp representation will cause highlight artifacts projected on the wall in this example.  The same artifact can be observed by comparing Fig. \ref{fig:teaser} (d) and (d.1) in the main paper, as in Fig. \ref{fig:teaser} (d) our lamp model projects wrong highlights on the wall behind. 

\vspace{-0.3cm}
\paragraph{Separation of shading and visibility} Our neural rendering framework separates visiibility ($\mathbf{S_j}$) and shading ($\mathbf{E_j}$) by assuming the direct shading $\mathbf{E_d}$ can be computed as
\begin{equation}
\mathbf{E_d} = \sum_{\mathbf{j}}\mathbf{S_jE_j}.
\label{eq:sup_directShg}
\end{equation}
There are two reasons we make this assumption. The first is that we hope to avoid checking the visibility for each ray in the rendering layer, which is too expensive and hard to be differentiable. The other is that we hope to introduce the shadow inpainting network $\mathbf{DShdNet}$, which can handle artifacts caused by occlusion boundaries robustly and is necessary when we render shadow from a mesh created from a single depth map. While Eq.~\eqref{eq:sup_directShg} works well for diffuse area lights, it may not work on directional light, where the visibility of each sampled ray should be considered separately. Fig.~\ref{fig:sup_shgFailed} compares $\mathbf{E_d}$ computed as in Eq.~\eqref{eq:sup_directShg} and the ground-truth direct shading. We can see that the ground-truth direct shading has more detailed highlight boundaries. 

\vspace{-0.3cm}
\paragraph{Missing geometry} While our shadow inpainting network can handle artifacts caused by occlusion boundaries, it cannot handle the case when occlusion causes a large region of geometry to be missing. Fig.~\ref{fig:sup_shdFailed} shows an example where rays go through the object because part of it is occluded and therefore cannot be reconstructed from a depth map. Some holistic single view mesh completion methods may help solve this problem, but this is beyond the scope of this paper. 

\vspace{-0.3cm}
\paragraph{One invisible lamp} While our one invisible lamp assumption works well practically, it can cause errors in specific regions. One example is shown in Fig. \ref{fig:teaser} (c) in the main paper. Compared to the real photo Fig. \ref{fig:teaser} (c.1), the lamp near the bed projects a wrong shadow on the wall because there are several small lamps on the ceiling lining against the wall in the real environment, while our method only predicts the major bright invisible lamp on top of the ceiling. 

\vspace{-0.3cm}
\paragraph{Future works} Currently, our framework can only handle a single image as the input. However, multi-view inputs can potentially lead to more complete and more accurate geometry reconstruction and more observation of the intensity distribution across the room. Therefore, it will be interesting to see how these multi-view inputs can help improve the indoor light editing results.

\bibliographystyle{splncs04}
\bibliography{egbib}
\end{document}